%% file: output.tex
\documentclass[11pt]{article}
\usepackage[margin=1in]{geometry}
\usepackage{adjustbox}
\usepackage{amssymb,amsmath}
\usepackage{graphicx}
\usepackage{hyperref}
\usepackage{authblk}
\usepackage{placeins}
\usepackage{subcaption}
\usepackage{tabularx}  % add this in your preamble
\usepackage[numbers,sort&compress]{natbib}
\usepackage{dsfont}

\hypersetup{
    colorlinks=true,
    linkcolor=blue,
    citecolor=blue,
    urlcolor=blue
}

\newcommand{\ie}{{\sl i.e.}}

\title{\textbf{The Curved Spacetime of Transformer Architectures}}

\author{
\textbf{Riccardo Di Sipio}$^{1}$ \quad
\textbf{Jairo Diaz-Rodriguez}$^{2}$\thanks{Supported by the Natural Sciences and Engineering Research Council of Canada (NSERC), through grant DGECR-2022-04531.} \quad
\textbf{Luis Serrano}$^{3}$\\
$^{1}$Dayforce, H.C.M. \quad
$^{2}$Department of Mathematics and Statistics, York University, Canada \quad $^{3}$Serrano Academy
}

\date{\today}

\begin{document}

\maketitle

%%%%%%%%%%

\begin{abstract}

\input{abstract} 
\end{abstract} %%%%%%%%%

\section{Introduction}
\label{sec:intro}

The astonishing performance of large language models (LLMs) has sparked renewed interest in the structures they form internally~\cite{brown2020language,bommasani2021opportunities}. The path to this moment began with Bengio’s neural probabilistic language model~\cite{bengio2003neural}, which demonstrated that predicting words in context could generate continuous vector embeddings, replacing symbolic representations with distributed ones. Word2Vec~\cite{word2vec}, GloVe~\cite{pennington2014glove}, and ELMo~\cite{peters2018deep} turned embeddings into the currency of natural language processing, while Transformers~\cite{vaswani2017attention} and successors such as BERT~\cite{devlin2019bert} and GPT~\cite{radford2018improving} transformed these static vectors into contextualized flows across layers.  

This echoes a trajectory in physics. Newton’s gravity invoked ``action at a distance", effective but conceptually unsettling. Einstein replaced it with geometry in his theory of General Relativity (GR): masses curve spacetime, and free trajectories follow geodesics~\cite{einstein1916grundlage,carroll2004spacetime}. Likewise, in language models, cosine similarity has served as a proxy for relational force between words: effective yet mechanisticly shallow. Recent work suggests that curved representational spaces may offer a deeper foundation~\cite{poincareglove2018,riemannformer2025,curveyourattention2025}.

%In this paper, we make the analogy explicit. Query–key interactions define a local metric, attention serves as a connection transporting embeddings across contexts, and stacked layers trace a discrete evolution of representations.

%In simpler terms, each Transformer layer induces its own curvature in the embedding space. Attention determines how tokens bend toward or away from one another, shaping the local geometry. As layers stack, this curvature evolves through the weights updated by gradient descent on the loss function, much like how, in general relativity, the distribution of mass and energy reshapes space and alters the trajectories of moving bodies over time. Thus, a token’s representation follows a curved path through successively adapting geometries, capturing progressively refined contextual meaning (Figure~\ref{fig:illustration}). These correspondences are summarized in Table~\ref{tab:gr-llm-analogy} which provides the conceptual map for the discussion that follows.

\begin{figure*}[ht]
    \centering
    \includegraphics[width=\textwidth]{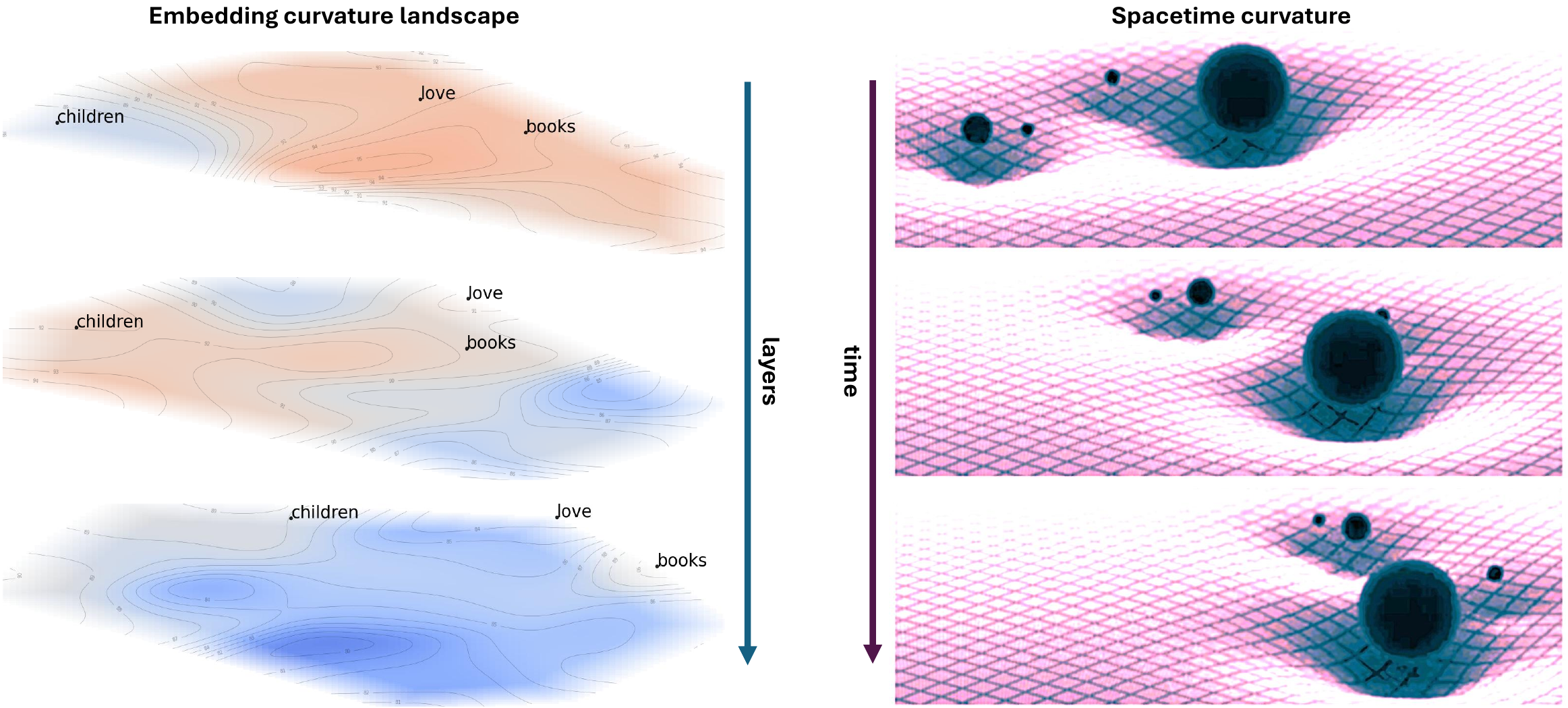}
    \caption{\small{
        Graphical analogy between spacetime curvature and embedding curvature.
On the right, spacetime curvature evolves over time according to the distribution of mass and energy. On the left, embedding curvature evolves across layers according to the learned weights.
    }}
    \label{fig:illustration}
\end{figure*}

%In this paper, we make the analogy explicit. Query–key interactions define a local metric, attention serves as a connection transporting embeddings across contexts, and stacked layers trace a discrete evolution of representations. These correspondences are summarized in Table~\ref{tab:gr-llm-analogy}, which provides the conceptual map for the discussion that follows.
 
\input{analogy_table}  
%This perspective does more than provide metaphor. By grounding Transformers in differential geometry, we clarify why embeddings evolve as they do and show how curvature emerges from training. The framework suggests concrete avenues for diagnostics, visualization, and even the design of architectures that explicitly embrace geometric principles. 

In this paper we make the analogy explicit. Query–key interactions define a local metric on representation space, attention acts as a connection that transports information across tokens, and stacked layers trace the discrete evolution of these representations through a curved manifold. In this view, each Transformer layer induces its own curvature: attention determines how tokens bend toward or away from one another, shaping the local geometry. As weights are updated through gradient descent, this curvature evolves, much like how the distribution of mass and energy reshapes spacetime curvature in GR. Thus, a token’s embedding follows a curved trajectory through successively adapting geometries, encoding progressively refined contextual meaning (Figure~\ref{fig:illustration}, Table~\ref{tab:gr-llm-analogy}).

This perspective moves beyond metaphor. Grounding Transformers in differential geometry offers an alternative way to think about how embeddings evolve and provides a means to test whether attention truly induces a curved representational geometry. To examine this, we introduce simple diagnostics that quantify curvature, \ie~a local measure based on turning angles between successive layer displacements and a global length-to-chord ratio capturing path elongation. These metrics allow us to ask whether the trajectories traced by token embeddings behave as if they inhabit a curved manifold, as our analogy predicts. Empirically, we observe systematic curvature that is unlikely to arise by chance from high dimensionality. Finally, we design an experiment inspired by Einstein’s 1919 eclipse test, which verified spacetime curvature through the deflection of starlight near the Sun. In our setting, contextual edits play the role of the gravitational source, and the resulting semantic deflections of token trajectories offer an embedding-space analogue of light bending, an empirical probe of the geometric analogy itself.

%\paragraph{Summary of findings. } 
%We propose a geometric interpretation of Transformer representations in which attention acts as a discrete connection over a curved semantic manifold.

%The bilinear form $g_{ij} = x_i^{\top} (W^{Q})^{\top} W^{K} x_j$ defines an effective metric that governs how representations interact and evolve across layers.

%From this, we derive a curvature proxy based on turning angles between successive layer displacements, capturing how trajectories of meaning bend as they propagate through depth.

%Empirical analyses (curvature heatmaps, layerwise profiles, and context-induced trajectory shifts) support this view, suggesting that large language models organize semantics along curved, context-dependent paths rather than linear projections through feature space.

\paragraph{Summary of contributions.}
(i) We propose a geometric interpretation of Transformer representations, where attention acts as a discrete connection transporting information across a curved semantic manifold, producing layerwise token trajectories shaped by learned geometry.
(ii) We test this analogy through experiments that visualize curvature, show it cannot be explained by dimensionality or chance, and include a contextual “deflection” test modeled after gravitational lensing, the phenomenon that revealed spacetime curvature, revealing consistent, meaning-dependent bends in embedding space.

%%%%%%%%%%%%%%%%%%%%%%
\section{Related Work}
\label{sec:related}

The geometry of embedding spaces has long been central to NLP. Early work by Bengio et al.~\cite{bengio2003neural} introduced neural probabilistic language models, opening the way to distributed representations. Subsequent approaches such as GloVe~\cite{pennington2014glove} and ELMo~\cite{peters2018deep} established embeddings as core components of modern systems~\cite{radford2018improving,devlin2019bert}, with cosine similarity in Euclidean space becoming the standard proxy for semantic closeness.

The limitations of flat geometry have since been recognized. Nickel and Kiela~\cite{poincareglove2018} showed that hyperbolic embeddings better capture hierarchical relations, motivating curved alternatives. More recently, Cho et al.~\cite{curveyourattention2025} introduced mixed-curvature Transformers with learnable sectional curvature, while Ji~\cite{riemannformer2025} proposed a principled Riemannian framework for attention based on parallel transport. Kratsios et al.~\cite{smalltransformers2024} further demonstrated that even compact Transformers can approximate universal metric embeddings. At a broader level, He et al.~\cite{nonEuclideanFoundation2024} argued for non-Euclidean geometries as a natural fit for linguistic and relational data.

These contributions examine curvature from empirical, architectural, and mathematical perspectives, but none explicitly connect it to the analogy with General Relativity. In particular, the view that semantic curvature arises dynamically from gradients in the training loss has not been developed. Our work extends this line of inquiry by making that connection explicit and by mapping Transformer mechanisms to differential geometric principles.

%%%%%%%%%%%%%%%%%%%%%%
\section{Mathematical Foundations of Transformer Geometry}
\label{sec:math_foundations}

Transformer models, introduced in \textit{Attention Is All You Need}~\cite{vaswani2017attention} and extended into BERT~\cite{devlin2019bert} and GPT~\cite{radford2018improving}, learn contextual embeddings that evolve across stacked layers.  
We interpret this process geometrically: queries and keys induce an effective metric, attention acts as a connection that governs transport, and layers correspond to discrete "time steps" along geodesic-like trajectories in semantic space. Tab. \ref{tab:geom_summary} summarizes the formal analogy.

\subsection{Query–Key–Value and Effective Metric}
\label{sec:metric}
Each token $x$ is projected into queries, keys, and values:
\begin{equation}
    Q = XW^Q, \quad K = XW^K, \quad V = XW^V,
\end{equation}
with learned matrices $W^Q, W^K, W^V \in \mathbb{R}^{d \times d_k}$.  
The dot product between queries and keys governs how  the model measures distances or angles between embeddings in the semantic space.

Hence, this bilinear form defines an \emph{effective metric}:
\begin{equation}
    g_{ij} := q_i^\top k_j = x_i^\top (W^Q)^\top W^K x_j.
\end{equation}
This effective metric is in general different from an identity matrix $\rm{diag}(1,1,\hdots,1)$~and thus sets the stage for defining curvature. Although $g_{ij}$ is not a Riemannian metric in the strict differential-geometric sense, being generally non-symmetric and not guaranteed positive-definite, it serves as the operational metric within the Transformer: all relational structure, attention, and contextual weighting derive from this kernel. This discrete construction does not replace the continuous Riemannian formalism but serves as its computational analogue, the way the Transformer operationalizes distance and orientation between semantic points in representation space.

\subsection{Attention as a Connection}
\label{sec:attention}
In Transformer networks, attention weights determine how much of each token contributes to updating another token’s representation. Scaled dot-product attention computes:
\begin{eqnarray}
    \text{Attention}(Q,K,V)_i = 
    \sum_j \hat{\alpha}_{ij} W^V x_j =
    \sum_j \hat{\alpha}_{ij} V_j, \\
    \hat{\alpha}_{ij} = \text{softmax}\!\left(\tfrac{Q_i \cdot K_j}{\sqrt{d_k}}\right).
\end{eqnarray}

\paragraph{}
The attention weights $\hat{\alpha}_{ij}$ determine how information from representation $v_j = W^Vx_j$ is transported to position $i$. In Riemannian geometry, the Christoffel symbols (connection coefficients) $\Gamma_{ij}^k$ describe how basis vectors change as one moves across the manifold, formally specifying how a vector is parallel-transported along coordinate direction $j$ from point $i$. 

\paragraph{}
In the Transformer, the indices $i$ and $j$ refer to tokens rather than coordinates, and the feature index $k$ is contracted within the value projection $W^V$. Thus, the attention operator $\hat{\alpha}_{ij} W^V$ acts as a discrete analogue of a connection: it defines how feature vectors are mixed and transported along edges of the token graph, rather than how coordinate bases change on a continuous manifold. Full derivations connecting attention to $\Gamma^k_{ij}$ appear in Appendix~\ref{app:intro_curved} and \ref{app:christoffel}. 

\subsection{Parallel Transport and Geodesics}
In differential geometry, parallel transport is defined by a connection $\Gamma^i_{jk}$ that determines how vectors change as they move along the manifold. The geodesic equation,

\begin{equation}
    \frac{d^2x^k}{dt^2} + \Gamma^k_{ij} 
    \frac{dx^i}{dt}\frac{dx^j}{dt} = 0
\end{equation}
describes curves whose tangent vectors are parallel-transported along themselves.

\paragraph{}
In a Transformer, attention defines an analogous discrete transport between tokens. The operator $\hat{\alpha}_{ij} W^V$ specifies how representations at point $j$ of the semantic manifold contributes to the updated representation at point $i$ within the same layer. The output projection $W^O$ acts as a transition maps between layers. Thus, the trajectory of a token embedding through Transformer layers approximates a discrete geodesic: each layer corresponds to a ``tick'' in semantic time.

\paragraph{}
By substituting the corresponding connection operator $\hat{\alpha}_{ij} W^V$~and replacing time derivatives with finite differences
into the geodesic equation, 
we obtain a layer-discretized evolution rule. 

Defining the hidden state at layer $\ell$ and the representation at the next layer $(\ell + 1)$:
\begin{eqnarray}
h_i^{(\ell)}   & = & \sum_j \hat{\alpha}^{(\ell)}_{ij}W_{(\ell)}^Vx^{(\ell)}_j, \\
x^{(\ell+1)}_i & = & x^{(\ell)}_i + W_{(\ell)}^O h_i^{(\ell)},
\end{eqnarray}

and the discrete velocity and acceleration:

\begin{eqnarray}
    \dot{x}^i_{(\ell)} &:=& x^i_{(\ell+1)} - x^i_{(\ell)}, \\
    \ddot{x}^i_{(\ell)} &:=& x^i_{(\ell+1)} - 2x^i_{(\ell)}, + x^i_{(\ell-1)}
\end{eqnarray}

then the discrete velocity $\dot{x}^i_{(\ell)}$ becomes:

\begin{eqnarray}
    \dot{x}^i_{(\ell)} = x^i_{(\ell+1)} - x^i_{(\ell)} =&
    W^O_{(\ell)} \sum_j \hat{\alpha}^{(\ell)}_{ij}W^Vx^{(\ell)}_j \\
    =& 
    \sum_j \left ( W^O_{(\ell)} \hat{\alpha}^{(\ell)}_{ij}W^V 
    \right ) x^{(\ell)}_j \\
    =& 
    \sum_j \Gamma_{ij}^{(\ell)} x_j^{(\ell)}
\end{eqnarray}

determining the discrete connection coefficients $\Gamma_{ij}^{(\ell)}$. Making use of a mid-point approximation (i.e $\Gamma_{ij}^{(\ell+1)}\approx\Gamma_{ij}^{(\ell)}$), the geodesic equation 
\begin{eqnarray}
    \ddot{x}^k_\ell + 
    \Gamma^k_{ij}(\ell)\,\dot{x}^i_\ell \dot{x}^j_\ell = 0,
\end{eqnarray}

becomes

\begin{equation}
x^i_{(\ell+1)} - 2 x^i_{(\ell)} + x^i_{(\ell-1)}
\sum_j \Gamma^{(\ell)}_{ij} 
\left (
x^i_{(\ell)} - x^j_{(\ell-1)}
\right ) = 0
\end{equation}

Thus each Transformer layer $\ell$ defines its own effective connection, 
and the forward pass can be seen as a discrete geodesic trajectory 
evolving through curved semantic space, in analogy with the ADM (Arnowitt–Deser–Misner) formalism of GR~\cite{arnowitt1962dynamics}.

\begin{table*}[ht]
\centering
\caption{A compact geometric summary of the Transformer layer.}
\renewcommand{\arraystretch}{1.25}
\begin{tabular}{@{}l l p{0.39\linewidth}@{}}
\hline
\textbf{Transformer component} & \textbf{Geometric analogue} & \textbf{Interpretation} \\
\hline
$W^{Q},\, W^{K}$ &
Metric tensor $g_{ij}$ &
Define the local geometry and pairwise inner products between token representations. \\

$q_i^\top k_j$ &
Covariant components of $g_{ij}$ &
Measure contextual similarity between tokens $i$ and $j$, analogous to metric components in a chosen basis. \\

$W^O_{(\ell)} \hat{\alpha}^{(\ell)}_{ij}W^V$ &
Connection coefficients $\Gamma^k_{ij}$ &
Act as discrete connection weights that determine how information flows or is transported from token $j$ to token $i$. \\

$W^{V} x_j$ &
Tangent--space vector $v_j$ &
Represent the quantity being transported, i.e the local semantic vector in the learned tangent basis. \\

$y_i = \sum_j \alpha_{ij} v_j$ &
Parallel transport of $v_j$ &
Performs a discrete parallel transport: aggregates and re-expresses the value vectors under the learned connection. \\

Stacked layers &
Geodesic integration &
Successive parallel transports across layers trace geodesic-like trajectories on the learned semantic manifold. \\
\hline
\end{tabular}
\label{tab:geom_summary}
\end{table*}

\subsection{A semantic least action principle.}
In general relativity, the Einstein–Hilbert action encodes the curvature of spacetime and its 
response to matter (Appendix~\ref{app:least_action} and \cite{landau1976mechanics, lanczos1949variational}). 
By analogy, we can define an effective action for language models in terms of 
their representational geometry and training objective:
\[
S_{\text{LM}} = \sum_{\ell=1}^L \Big( g_{ij}^{(\ell)} \, \dot{x}^i_{(\ell)} \dot{x}^j_{(\ell)} - \mathcal{L}_{\text{train}}(x_{(\ell)}; \theta) \Big),
\]
where \( g_{ij}^{(\ell)} \) is the effective metric induced by query–key interactions at layer $\ell$, 
\( \dot{x}^i_\ell \) denotes the discrete update of token embeddings across layers, and 
\( \mathcal{L}_{\text{train}} \) is the loss functional that couples geometry to data. A full derivation is outlined in App.~\ref{app:least_action_semantic}.

\paragraph{}
Backpropagation can then be interpreted as the variational procedure that extremizes this action, 
ensuring that forward trajectories $x_\ell$ evolve along paths that minimize training loss. 
In this analogy, the dataset itself plays the role of ``mass-energy'': 
its distribution acts as the source term shaping curvature in embedding space. 
Thus, just as in general relativity matter tells spacetime how to curve, 
data tells language models how to bend their representational manifold, 
and the resulting dynamics follow a least-action principle.

\paragraph{}
However, some words of caution are warranted in this context. In our setting, the layerwise activations $x_\ell$ are not independent coordinates that can be varied continuously, but functions of the model parameters $\theta$. In the representation space, there exist paths $x_\ell$ that the model will never traverse, because no combination of weights $\theta$ generates them. Accordingly, the functional $S_{LM}$ should be understood as an action over parameter-induced trajectories $x_\ell(\theta)$ in representation space, rather than over arbitrary paths as it is assumed in Riemann geometry. Extremizing $S_{LM}$ with respect to $\theta$ corresponds to adjusting the weights so as to minimize both the ``kinetic'' cost of layer-to-layer changes (measured by the effective metric $g_{ij}$) and the ``potential'' cost given by the training loss.
In this sense, backpropagation computes the gradient of an action functional defined on the weight manifold, with $g^{(l)}_{ij}$ acting as a {\sl pull-back metric} linking parameter changes to geometric motion in the embedding space.

\subsection{Multi-Head Attention as an Atlas of Charts}
Each attention head defines its own $(W^Q, W^K, W^V)$ projections, effectively a separate chart on the manifold. Just as multiple coordinate charts form an atlas in differential geometry to avoid singularities, multi-head attention provides overlapping local views. The output projection $W^O$ acts as a transition map, integrating these views into a global representation.

\subsection{Theoretical summary and experimental observables}

The framework developed above treats attention as a discrete connection and defines curvature as the deviation of token trajectories from locally geodesic flow.

In practice, the true curvature of the representation manifold cannot be observed directly, since the manifold itself is implicit in the model’s parameters.
Instead, we estimate it through observable proxies derived from embedding trajectories: the turning angle between successive layer displacements (local curvature) and the length-to-chord ratio measuring global deviation.
These metrics provide experimentally accessible traces of the underlying geometry.

The goal of the experiments is therefore not to measure curvature directly but to test whether the observed embedding dynamics require a curved geometric explanation.
If the trajectories of representations follow the same relational patterns that characterize geodesic deviation, curvature becomes the natural framework to describe them.
In this sense, the curvature proxies serve as indirect evidence of the manifold’s shape, translating abstract geometric principles into measurable effects within Transformer activations.

\section{Experiments}

%%%%%%%%%%%%%%%%%%%%%%
In this section, we present the results of three experiments aiming at demonstrating the curvature of the embeddings.

To quantify the curvature of token trajectories in representation space we introduce the following concept:

\paragraph{Local curvature (turning angle).}  
\label{sec:geometric-trajectory}
Given embeddings $x_{i-1}$, $x_{i}$, and $x_{i+1}$ of the same token at consecutive layers, we define step vectors $\Delta_i = x_i - x_{i-1}$ and $\Delta_{i+1} = x_{i+1} - x_{i}$. 
The turning angle $\theta_i$ is:
\begin{equation}
    \theta_i = \arccos
    \left(
    \frac{\Delta_i \cdot \Delta_{i+1}}
         {\lVert \Delta_i \rVert \, \lVert \Delta_{i+1} \rVert}
    \right).
\end{equation}
In differential geometry, curvature is formally defined as the rate of change of a tangent vector with respect to arc length, $\kappa(s) = \lVert dT/ds \rVert$.  
Here, the tangent at layer $i$ is approximated by $\Delta_i$, and $\theta_i$ serves as a discrete proxy for local curvature: 
small angles indicate nearly straight trajectories, while large angles signal sharp bends.  
In high-dimensional embedding spaces, random vectors are almost orthogonal ($\approx 90^\circ$). 
Thus, deviations below $80^\circ$ indicate alignment toward a straight path (flat angles), while values above $100^\circ$ reflect strong reorientations (sharp angles).

\subsection{Example of  the Curvature Landscape of a Paragraph}
\label{sec:curvature_landscapes}

We first illustrate how an entire paragraph evolves through the Transformer stack. For this experiment, we use the following text (69 words) as input to the  BERT transformer model~\cite{devlin2019bert}~(12 layers, hidden size 768, 12 attention heads, ~110M parameters): 

\begin{quote}
The library was filled with books and stories. Students gathered in the hall to read, while parents encouraged children to discover new worlds through pictures and exams. The community saw the place as more than a building: it was a meeting point, open to everyone, where older generations shared their knowledge with the young, and workshops supported writing skills.
\end{quote}

We track each token embedding across layers, projected to two principal components via PCA, and analyze for local curvature using the turning angle measure defined in Sec.~\ref{sec:geometric-trajectory}. 
This yields a geometric ``landscape'' where tokens act like pebbles carried downstream: some drift smoothly, while others bend sharply in response to contextual forces.

\begin{figure}[h!]
    \centering
    \includegraphics[width=0.4\textwidth]{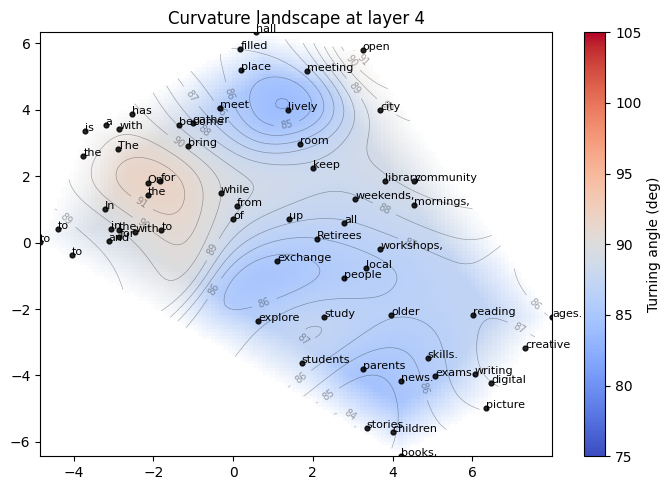}
    \includegraphics[width=0.4\textwidth]{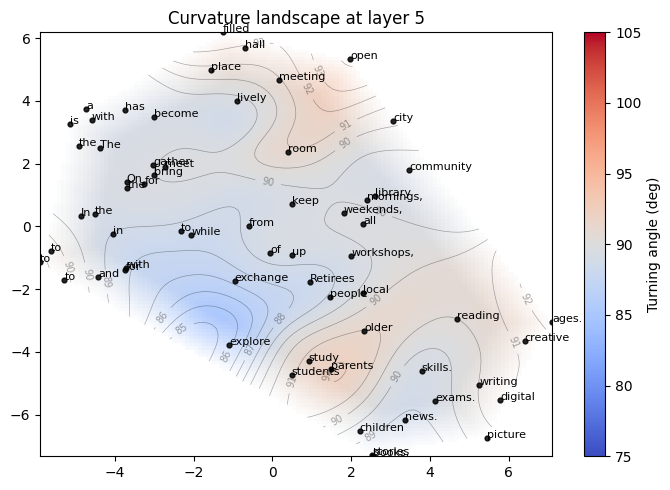}
    \includegraphics[width=0.4\textwidth]{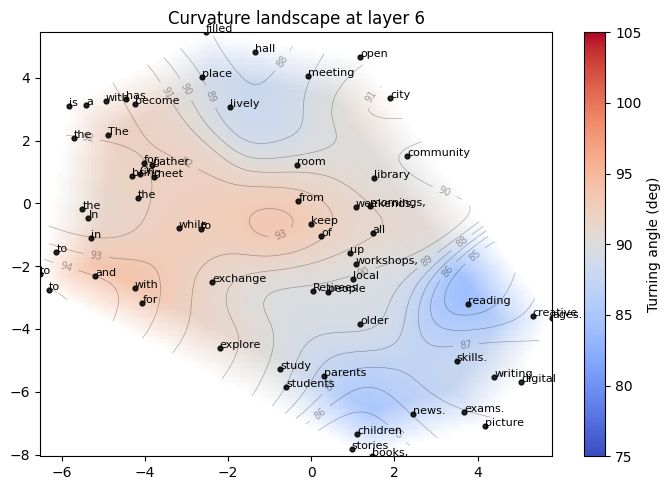}
    \includegraphics[width=0.4\textwidth]{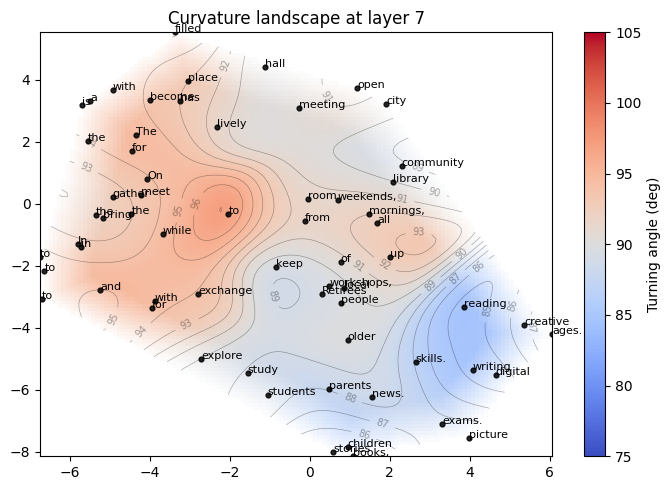}
    \includegraphics[width=0.4\textwidth]{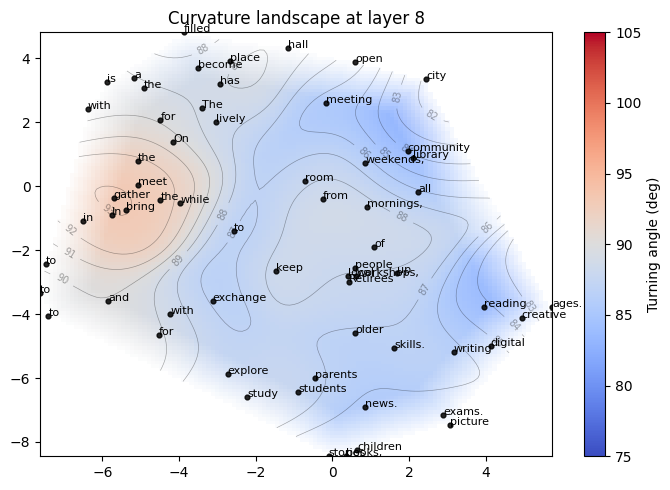}
    \includegraphics[width=0.4\textwidth]{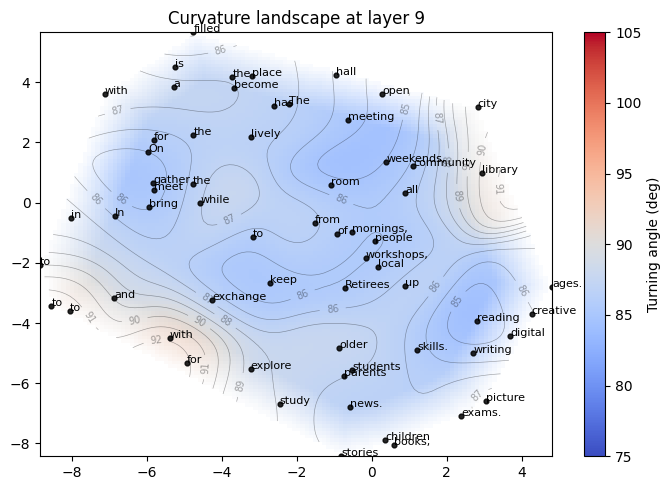}
    \caption{\small{
        2D curvature landscape for the same paragraph at six consecutive layers.  
        Token positions are projected onto a PCA plane, with color encoding turning angle:  
        blue areas indicate straighter motion ($<90^\circ$), and red areas indicate sharper bending ($>90^\circ$).  
        This heatmap highlights regions of high contextual curvature where embeddings are more actively reshaped by attention.
    }}
    \label{fig:exp4-heatmap}
\end{figure}

\paragraph{Curvature Heatmaps.}
Figure~\ref{fig:exp4-heatmap} shows the curvature proxy as a heatmap: token positions are shown in PCA space, with color denoting curvature relative to an average turning angle of 90 degrees.
Interestingly, the highest curvature peak corresponds not to a semantically heavy word, but to the function word \textit{to}. This suggests that even seemingly lightweight tokens can act as critical junctions in the representational flow, depending on context. 
Other high-curvature regions are observed near \textit{books}, \textit{parents}, and \textit{filled}, while words such as \textit{the}, \textit{is}, and \textit{of} remain closer to flat trajectories. The evolution of curvature across layers progresses smoothly, indicating an underlying continuous process that is being captures in ``snapshots'' at each time slice.

\paragraph{Semantic clustering.}  
The projection also reveals two clear semantic clusters. 
One group (\textit{parents}, \textit{students}, \textit{children}, \textit{books}, \textit{stories}, \textit{exams}, \textit{pictures}, \textit{older}) revolves around education and learning.  
Another group (\textit{hall}, \textit{place}, \textit{filled}, \textit{meeting}, \textit{open}, \textit{city}, \textit{library}, \textit{community}) captures spaces of gathering and collective activity.  
These clusters highlight how curvature is not random but organizes tokens into coherent neighborhoods that reflect semantic themes.

\begin{figure}[h!]
    \centering
    \includegraphics[width=0.5\textwidth]{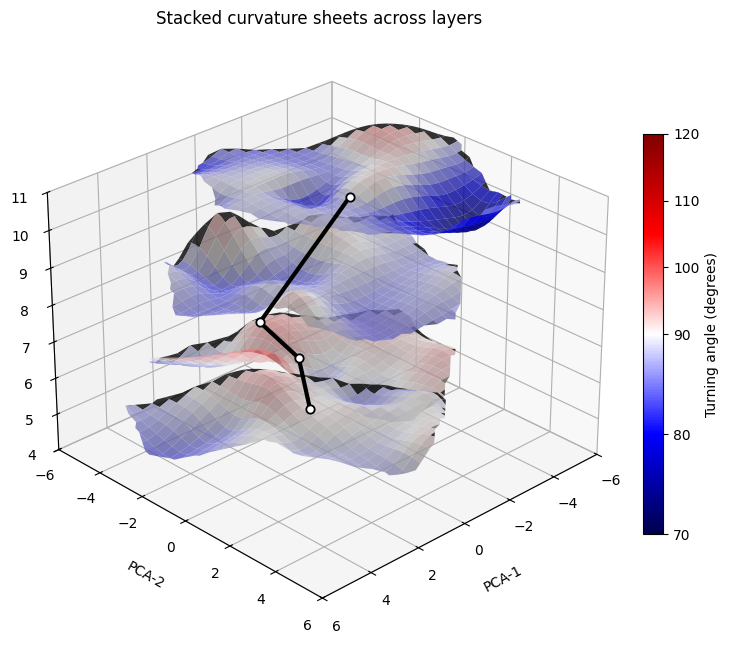}
    \caption{\small{
        Foliated heatmap visualization of token embeddings across Transformer layers.
        Each wavy sheet corresponds to one layer, colored by the local turning angle
        (blue: $<90^\circ$, red: $>90^\circ$). The vertical axis represents discrete
        ``semantic time'' as the paragraph is processed layer by layer. The trajectory
        of tokens can be traced across sheets, illustrating how local curvature evolves
        throughout the model.
    }}
    \label{fig:paragraph_foliated_heatmap}
\end{figure}

\paragraph{}
Beyond surface projections, it is also instructive to view the evolution of embeddings
as a \emph{stacked foliation} across layers. In this view, each sheet corresponds to
a Transformer layer, and the stack as a whole captures how semantic states propagate
through depth. As shown in Fig.~\ref{fig:paragraph_foliated_heatmap}, tokens trace
discretized trajectories through ``semantic time.'' The contrast between red (sharp
turns) and blue (smooth transitions) reveals variations in local curvature and allows
trajectories such as that of \textit{books} to be followed across layers, in analogy
to the foliation of spacetime in general relativity.

\paragraph{Interpretation.}  
These visualizations illustrate that curvature is not uniformly distributed across tokens. 
Function words, often assumed to be semantically light, can generate sharp local bends, while content words produce smoother but still context-sensitive peaks. 
The landscape perspective therefore provides an intuitive, geometry-based tool for inspecting how context reshapes language representations, revealing unexpected contributors to semantic curvature.

\subsection{Curvature Analysis Across Layers}\label{exp:curvature}

We now study how token representations evolve across Transformer layers by quantifying their geometric curvature in embedding space. First, we define global curvature:

\paragraph{Global curvature (path length).}  
For embeddings $x_0, x_1, \ldots, x_N$, the trajectory length is the polyline arc:
$
    L = \sum_{i=1}^N \lVert x_i - x_{i-1} \rVert.
$
This quantity captures the accumulated displacement as a token evolves across layers.  
In Riemannian geometry, the length of a curve $\gamma(s)$ is obtained by integrating the norm of its tangent vector, $L = \int \lVert \dot{\gamma}(s) \rVert ds$.  
Our discrete sum is a layer-wise analogue: if the manifold were flat and the update directions aligned, $L$ would closely match the direct endpoint distance $\lVert x_N - x_0 \rVert$.  
Longer paths relative to this displacement imply higher curvature, since the representation bends and reorients instead of following a straight geodesic through embedding space. To measure this relation we focus on the length-to-cord ratio
\[
R = \frac{\sum_i \|\Delta_i\|}{\|x_N - x_0\|}.
\]

\paragraph{Setup. }In this experiment we analyze multiple token embedding trajectories through its sequence of turning angles $\theta_i$ and its overall length-to-chord ratio $R$.
To test whether such directional changes reflect structured contextual transformations rather than random drift, we construct a \emph{null model}. 
For each trajectory, the step lengths $\{\|\Delta_i\|\}$ are preserved, while directions are replaced by random unit vectors, generating $1000$ synthetic trajectories per word. 
This control accounts for the anisotropy and scaling inherent in Transformer embeddings; if layer transitions were random, the transformer and null trajectories would be similar. See a detailed explanation on the construction of these null trajectories in Appendix~\ref{app:construction}.
We apply this analysis in two settings:
\begin{itemize} 
\item First, we run a transformer model through each of 100 random sentences from the High-Quality English Sentences dataset \citep{agentlans_high_quality_english_sentences_2024}, separately.
From each, one noun or verb is selected via \texttt{spaCy}, and its embedding trajectories are extracted. 
\item Second, we apply the same procedure to the first paragraph of \emph{One Hundred Years of Solitude} \citep{marquez1967}, treating each word as an individual trajectory within a shared paragraph context. 
\end{itemize}
The paragraph setting tests whether curvature persists under strong inter-word dependencies, where contextualization evolves continuously across the discourse.

\paragraph{Results. } Table~\ref{tab:sentences} reports the total number of flat ($\theta<80^\circ$) and sharp ($\theta>100^\circ$) angles and the mean length–to–chord ratio across all trajectories in the first experiment, for both the observed Transformer trajectories and those generated under the null model, across multiple embedding architectures (Results for the second experiment are analogous and deferrd to Table~\ref{tab:paragraph} in Appendix).
A larger number of flat and sharp angles indicates trajectories that follow purposefully directed paths rather than random high-dimensional artifacts, while a higher length–to–chord ratio ($R$) reflects stronger curvature, as embeddings bend and reorient instead of following straight geodesics through representation space.  
For all models, Transformer trajectories display markedly higher counts and $R$ values than their null counterparts, providing exploratory evidence of embedding-space curvature.  
Appendix~\ref{app:null} presents the confirmatory statistical analysis, showing through pooled and paired hypothesis tests that these effects are highly significant and not attributable to chance.  
Together, these results demonstrate that contextual embeddings evolve along structured, non-linear trajectories, alternating between phases of alignment and sharp reorientation—offering geometric evidence that contextualization in Transformers arises from directional bending of embeddings across layers rather than linear or random translation in feature space.

\begin{table*}[h]
\caption{Curvature metrics of 100 embedding trajectories across Transformer layers on multiple models (see Appendix~\ref{app:transformers}).
For each model, we report counts of flat ($\theta<80^\circ$) and sharp ($\theta>100^\circ$) angles, and the average length-to-chord ratio $R$, compared to a randomized null model sampled 1000 times.}
\centering
\small
\renewcommand{\arraystretch}{1.3}

\begin{tabular}{lllrrrcrrr}
\hline
 &  &  & \multicolumn{3}{c}{\textbf{Transformers}} &  & \multicolumn{3}{c}{\textbf{Null model}} \\
\cline{4-6} \cline{8-10}
\textbf{Size} & \textbf{Model} & \textbf{Angles} & \textbf{flat} & \textbf{sharp} & \textbf{Average $R$} &&  \textbf{flat} & \textbf{sharp} & \textbf{Average $R$} \\
\hline
small & DistilBERT & 500 & 89 & 70 & 3.29 &  & 0.00 & 0.00 & 2.39 \\
 & DistilRoBERTa & 500 & 193 & 0 & 3.73 &  & 0.00 & 0.00 & 2.28 \\
 & MiniLM & 500 & 25 & 57 & 3.47 &  & 0.15 & 0.15 & 2.40 \\
 \hline
base & BERT & 1100 & 25 & 20 & 5.29 &  & 0.00 & 0.00 & 3.42 \\
 & RoBERTa & 1100 & 123 & 53 & 6.03 &  & 0.00 & 0.00 & 3.24 \\
 & MiniLM & 1100 & 75 & 53 & 5.72 &  & 0.34 & 0.34 & 3.43 \\
 & DeBERTa & 1100 & 21 & 236 & 2.71 &  & 0.00 & 0.00 & 2.63 \\
 & ALBERT & 1100 & 44 & 53 & 4.30 &  & 0.00 & 0.00 & 3.23 \\
 \hline
large & BERT & 2300 & 527 & 106 & 10.07 &  & 0.00 & 0.00 & 4.72 \\
 & RoBERTa & 2300 & 563 & 23 & 5.05 &  & 0.00 & 0.00 & 4.62 \\
 & DeBERTa & 2300 & 142 & 1 & 6.45 &  & 0.00 & 0.00 & 4.79 \\
 & ALBERT & 2300 & 428 & 227 & 6.23 &  & 0.00 & 0.00 & 4.52 \\
\hline
\end{tabular}
\label{tab:sentences}
\end{table*}

\subsection{Context-Induced ``Gravitational Lensing'' of Meaning}\label{sec:gravitational_lensing}

\begin{figure*}[htb]
  \centering
  \begin{subfigure}[t]{0.32\textwidth}
    \centering
    \includegraphics[width=\linewidth]{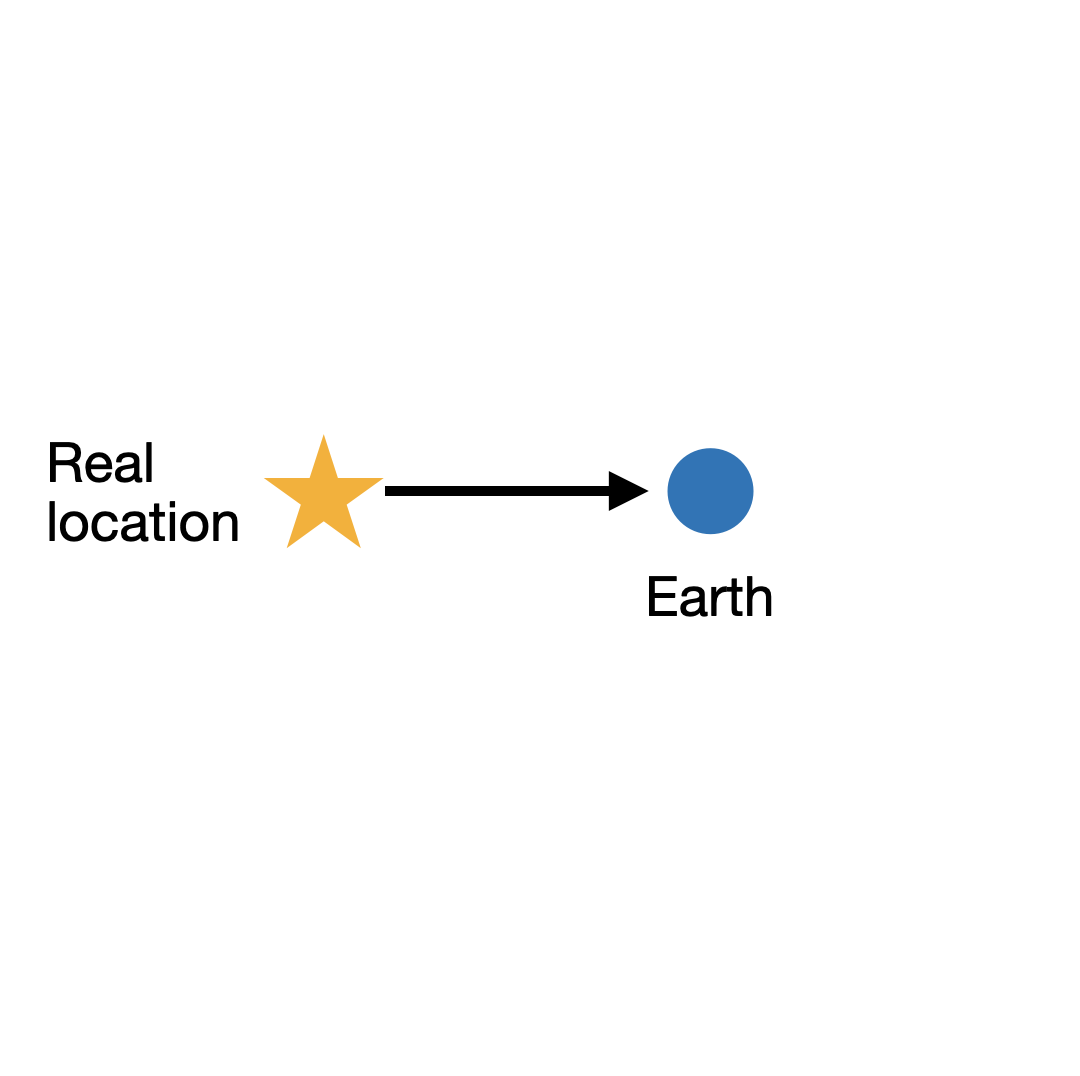}
    \caption{Light travels in a straight line without curvature.}
    \label{fig:starlight_a}
  \end{subfigure}
  \hfill
  \begin{subfigure}[t]{0.32\textwidth}
    \centering
    \includegraphics[width=\linewidth]{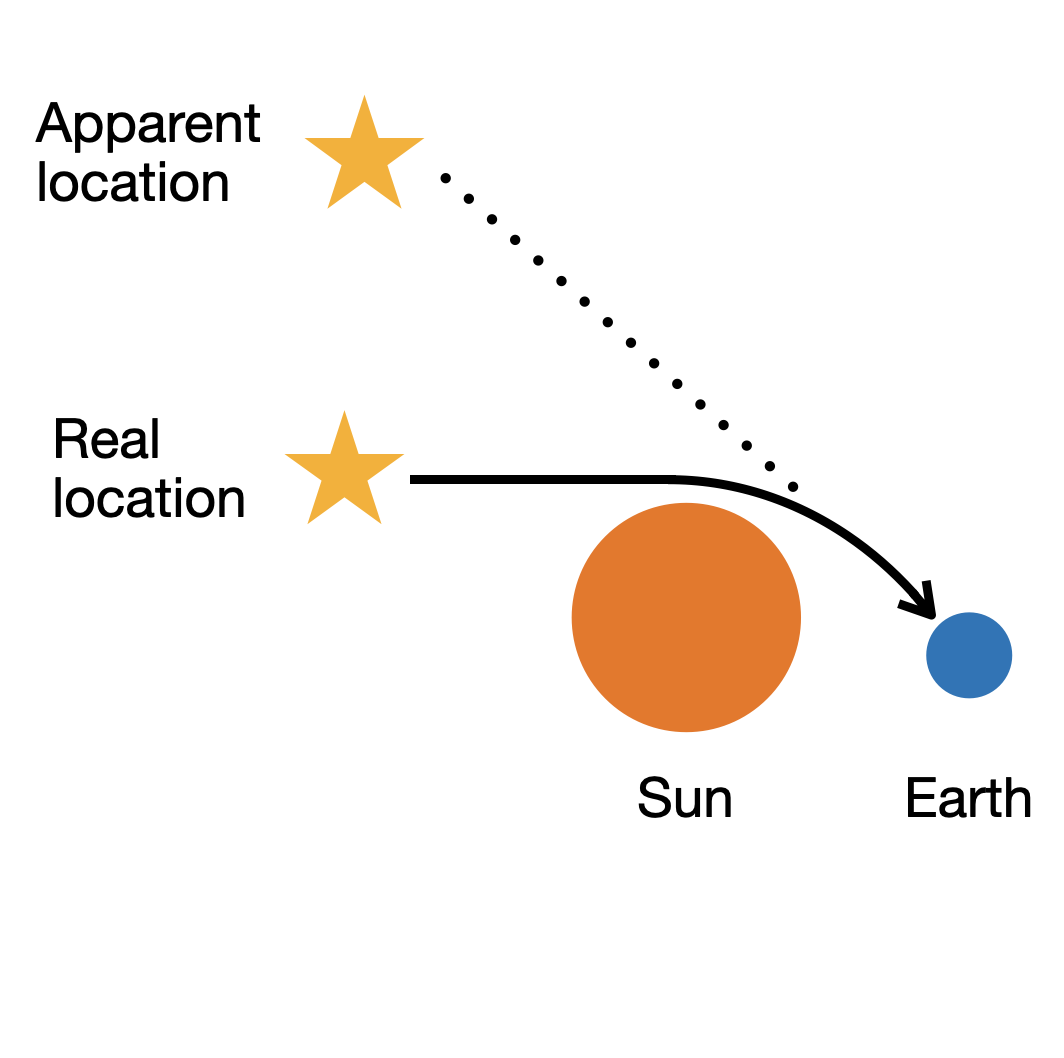}
    \caption{Deflection near the Sun due to spacetime curvature.}
    \label{fig:starlight_b}
  \end{subfigure}
  \hfill
  \begin{subfigure}[t]{0.32\textwidth}
    \centering
    \includegraphics[width=\linewidth]{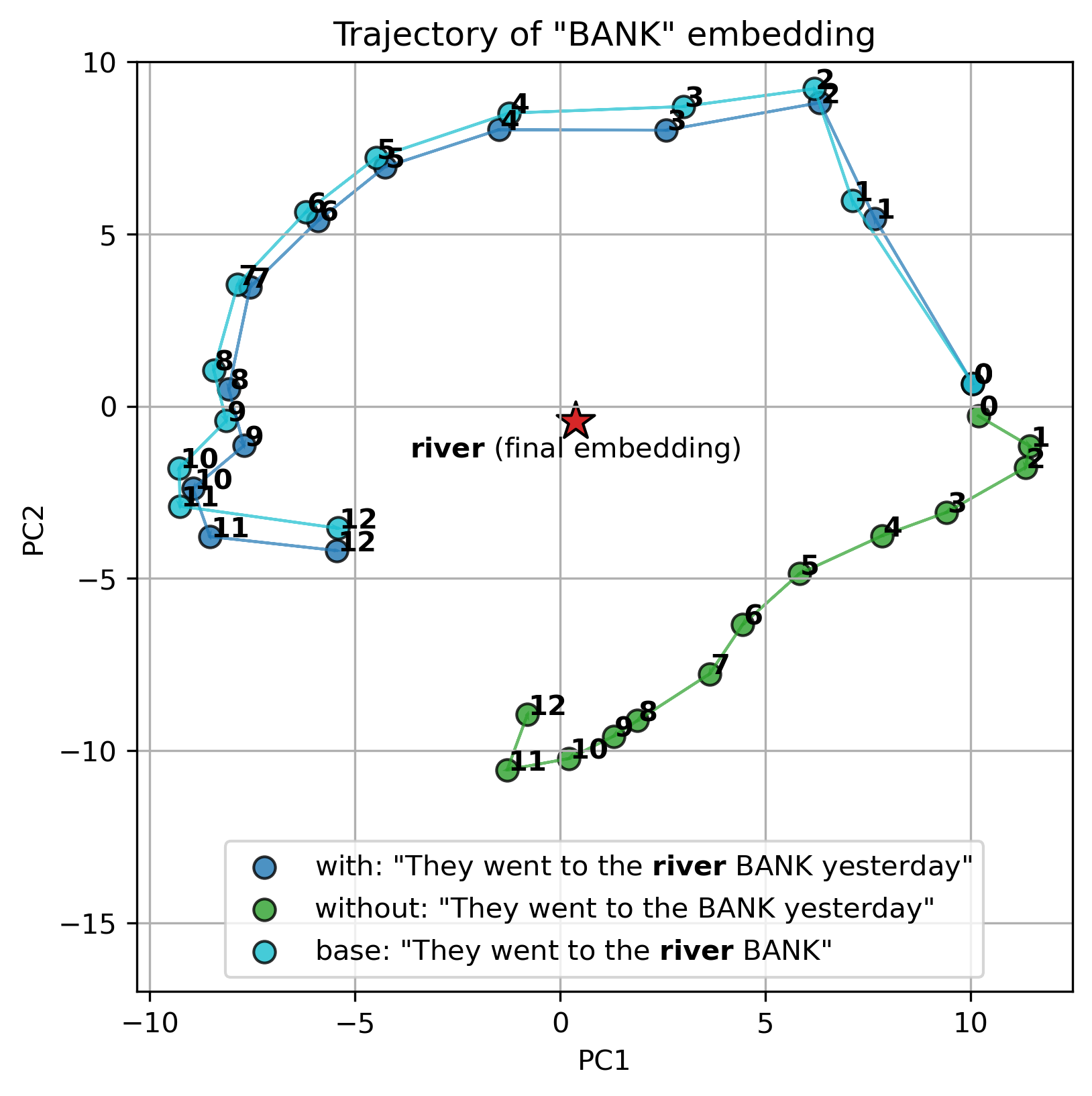}
    \caption{Trajectory of “BANK” embedding in a Transformer.}
    \label{fig:starlight_c}
  \end{subfigure}
  \caption{\small
    Analogy between relativistic curvature and embedding curvature.  
    (a) Without curvature, light travels in a straight line.  
    (b) Near the Sun, spacetime curvature bends light, producing an apparent shift.  
    (c) In language models, the token “BANK” follows a curved trajectory in representation space depending on its semantic context.
  }
  \label{fig:starlight}
\end{figure*}

Our last experiment is inspired on Einstein's classic 1919 solar eclipse experiment. A central claim of General Relativity is that mass curves spacetime, and that this curvature can be empirically detected by measuring how light rays bend in the presence of a massive object. 
The classic 1919 solar eclipse experiment tested this: starlight passing near the sun appeared displaced relative to its expected straight-line trajectory, because the sun's mass curved spacetime and deflected the path of the light (see the illustration in Fig. \ref{fig:starlight}).
Operationally, the test compares two trajectories of the \emph{same} underlying signal---(i) with the massive body in view and (ii) without it---and attributes any systematic deflection to curvature induced by mass.

We adopt this logic in representation space in this. 
We treat an ambiguous token (e.g., ``bank'') as the ``light ray,'' and a contextual disambiguator (e.g., ``river'') as the ``mass.'' 
If context (``river'') exerts semantic force on the token (``bank''), it should not only change the token's final embedding, it should bend its trajectory across layers. 
To test this, we construct controlled triples of sentences:
\begin{itemize}
    \item \textbf{with:} a sentence in which a disambiguating modifier forces a specific sense of the target word (e.g., ``They went to the \emph{river} bank yesterday.'').
    \item \textbf{without:} the same sentence with that modifier removed (``They went to the bank yesterday.''), allowing the target word to revert to an alternative sense.
    \item \textbf{base:} a control edit in which we remove a token that is not the disambiguator (``They went to the river bank.''), to account for generic perturbation effects unrelated to the meaning of the target word.
\end{itemize}
Here the disambiguator (``river'') plays the role of the gravitational source: if it truly ``curves'' representation space, then the trajectory of ``bank'' in the \texttt{with} sentence should deviate from its trajectory in the \texttt{without} sentence in a way that exceeds the deviation observed in the \texttt{base} condition.

\paragraph{Setup.} For each sentence $s \in \{\texttt{with}, \texttt{without}, \texttt{base}\}$, we extract the layerwise embedding trajectory of the target token, $\{x^{(s)}_0, \dots, x^{(s)}_N\}$. 
We then compare trajectories between sentences within each triple. 
Given two trajectories $a$ and $b$ (e.g., \texttt{with} vs.\ \texttt{without}), we report four quantities:
\begin{enumerate}
    \item \textbf{Final-layer separation.} 
    The cosine distance between the final-layer embeddings,
    \begin{equation}
        d_{\mathrm{final}}(a,b) 
        = 1 - \frac{\langle x^{(a)}_N, x^{(b)}_N \rangle}{\|x^{(a)}_N\| \, \|x^{(b)}_N\|},
    \end{equation}
    which tests whether the two contexts drive the target token to distinct semantic endpoints.
    \item \textbf{Layerwise separation.}
    The mean Euclidean distance between trajectories across depth,
    \begin{equation}
        d_{\mathrm{layer}}(a,b) 
        = \frac{1}{N+1} \sum_{i=0}^{N} \bigl\| x^{(a)}_i - x^{(b)}_i \bigr\|,
    \end{equation}
    which reveals how persistently the trajectories diverge.
    \item \textbf{Curvature divergence.}
    The mismatch in how the token bends across layers,
    \begin{equation}
        \Delta_{\mathrm{curv}}(a,b)
        = \frac{1}{N} \sum_{i} \Biggl( 1 
        - \frac{\langle \Gamma^{(a)}_i, \Gamma^{(b)}_i \rangle}
               {\|\Gamma^{(a)}_i\| \, \|\Gamma^{(b)}_i\|} \Biggr),
    \end{equation}
   where $\Gamma^{(s)}_i = \Delta^{(s)}_{i+1} - \Delta^{(s)}_{i}$. Larger values indicate that the two trajectories do not merely end in different places; they curve differently along the way.
    \item \textbf{Turning-angle gap.}
    The mean absolute difference in turning angles across layers,
    \begin{equation}
        \Delta_{\theta}(a,b) 
        = \frac{1}{N} \sum_{i} 
        \bigl| \theta^{(a)}_i - \theta^{(b)}_i \bigr|.
    \end{equation}
    This measures local reorientation differences.
\end{enumerate}

We compute these metrics for all three pairwise comparisons within each triple (\texttt{with} vs.\ \texttt{without}, \texttt{with} vs.\ \texttt{base}, \texttt{base} vs.\ \texttt{without}). 
The key test is whether \texttt{with} vs.\ \texttt{without} exhibits larger quantities than \texttt{with} vs.\ \texttt{base}. 
If so, then the contextual token (e.g., ``river'') is not simply adding information downstream at the final layer; it is actively bending the trajectory of the ambiguous token throughout the network.

\begin{figure*}[tb]
    \centering
    \includegraphics[width=\textwidth]{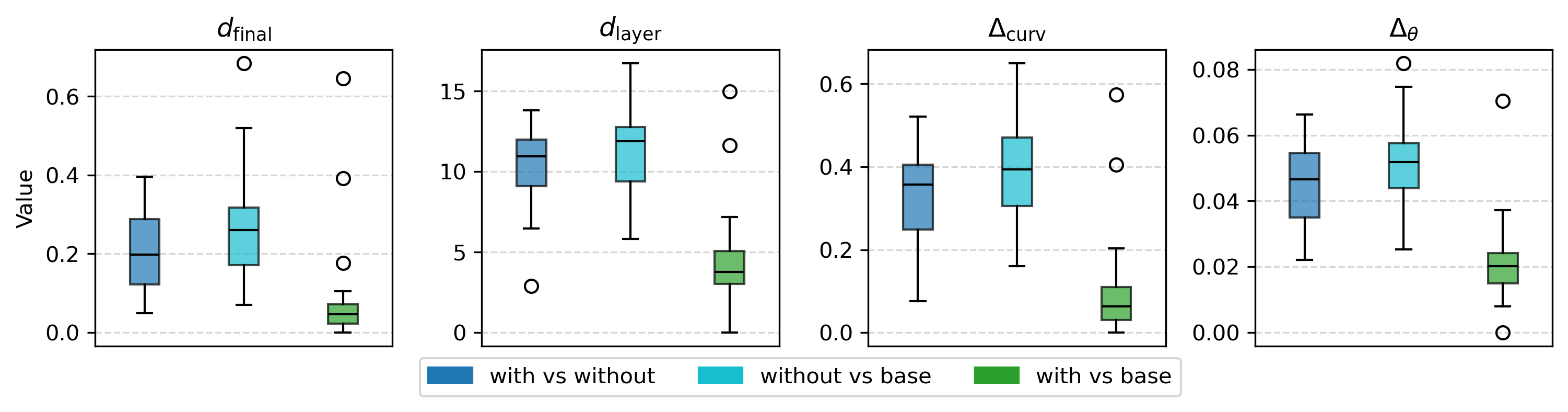}
    \caption{\small{
       Trajectory divergence results for \texttt{bert-base-uncased}. 
Each subplot reports one of the four divergence metrics computed across 50 sentence triples. 
Boxplots show pairwise comparisons between sentence variants (\texttt{with} vs.~\texttt{without}, \texttt{without} vs.~\texttt{base}, \texttt{with} vs.~\texttt{base}); 
higher values denote stronger divergence between embedding trajectories.
    }}
    \label{fig:bertsun}
\end{figure*}

\paragraph{Results.}
Figure~\ref{fig:bertsun} summarizes the results for BERT \citep{devlin2019bert}.
Each of the four subplots corresponds to one trajectory-divergence metric, computed across 50 sentence triples.
Within each subplot, the three boxplots represent pairwise comparisons between sentence variants: \texttt{with}~vs.~\texttt{without}, \texttt{without}~vs.~\texttt{base}, and \texttt{with}~vs.~\texttt{base}.
Higher values indicate greater divergence between the corresponding embedding trajectories.

We observe that divergences are consistently larger for \texttt{with}~vs.~\texttt{without} and \texttt{without}~vs.~\texttt{base} than for \texttt{with}~vs.~\texttt{base}, suggesting that removing the disambiguating token alters the trajectory more strongly than adding it.
This trend holds across both positional measures ($d_{\mathrm{layer}}$, $d_{\mathrm{final}}$) and geometric measures ($\Delta_{\mathrm{curv}}$, $\Delta_{\theta}$).

In other words, the presence of the disambiguator systematically changes not only where a representation ends, but how it moves through the network’s depth.
This serves as a causal probe of curvature: contextual tokens act like localized mass, bending nearby embedding trajectories in a manner reminiscent of gravitational lensing.
The addition or removal of a single semantically loaded word thus induces measurable deflections in the trajectory of an adjacent ambiguous token.
Complete results across multiple embedding models appear in Table~\ref{tab:sun} in the Appendix.

%%%%%%%%%%%%%%%%%%%%%%
\section{Discussion}
\label{sec:discussion}
We have proposed that language models, particularly Transformers, operate in a latent geometric space whose curvature reflects the structure of language itself (Tab.~\ref{tab:gr-llm-analogy}). This geometry is not fixed a priori, but emerges during training from the distribution of data and the gradients of the loss function.

A key idea is to interpret each Transformer block as a discrete ``tick'' in a semantic time-evolution. Each layer updates token representations via attention dynamics and learned weights, analogous to a time step in a discretized physical system. This invites comparison with the Hamiltonian formulation of GR, where spacetime is ``foliated'' into successive slices indexed by a time parameter~\cite{arnowitt1962dynamics}. Similarly, the sequence of Transformer layers may be viewed as a discrete foliation of semantic space: each slice reshapes the manifold while preserving coherence across layers.

This perspective suggests that interpretability might benefit from geometric tools. If each layer applies a differential transformation akin to parallel transport, then methods from differential geometry may help visualize and analyze how representations evolve. It also raises the possibility that depth should not be treated as an arbitrary hyperparameter, but as the number of steps in a meaningful dynamical process, perhaps governed by deeper variational principles.

We note a limitation of the analogy. In General Relativity, curvature and matter interact dynamically (back-reaction), whereas Transformer inference unfolds in a static geometry: the $W^Q$ and $W^K$ matrices define the manifold but remain fixed once trained. A natural direction for future work is to explore architectures where these matrices adapt during inference, introducing a form of geometric feedback. Such models could allow richer semantic generalization and dynamic contextualization.

A final remark is that the relativity of this framework lies not in the existence of curvature per se, but in its observability. We do not see the global structure of the semantic manifold; rather, we infer curvature locally, along the specific sentences and contexts we probe. In this sense, the analogy mirrors GR, where curvature is detected only through the motion of particles and light along particular paths.

%%%%%%%%%%%%%%%%%%%%%%
\section{Conclusions}
\label{sec:conclusions}

We have outlined a geometric interpretation of Transformer-based language models, framing attention and query–key–value dynamics as connections that guide the transport of token representations through a curved semantic space. In this view, language understanding is not the result of static similarity but of trajectories shaped by context and curvature.

While prior work has hinted at non-Euclidean structures in embeddings, our framework connects these observations to the mathematical machinery of differential geometry which sits at the base of Einstein's theory of General Relativity, offering a unifying perspective. This suggests new directions for model design: architectures that move beyond flat space and perhaps incorporate adaptive, curvature-aware mechanisms.

Ultimately, understanding in language models can be seen not as a lookup, but as a journey: a geodesic traced through a manifold bent by data, training, and context.

\paragraph{Reproducible research.} The source code to replicate all experiments in this paper can be found on GitHub: \href{https://github.com/rdisipio/llm-curvature}{https://github.com/rdisipio/llm-curvature}

\input{acknowledgments}

\bibliographystyle{apalike}
\bibliography{bibliography}

%%%%%%%%%%%%%%%%%%%%%%%%%%%%%%%%%%%%%%%%%%%%%%%%%%%%%%%%%%%%%%%%%%%%%%%%%%%%%%%
%%%%%%%%%%%%%%%%%%%%%%%%%%%%%%%%%%%%%%%%%%%%%%%%%%%%%%%%%%%%%%%%%%%%%%%%%%%%%%%
% APPENDIX
%%%%%%%%%%%%%%%%%%%%%%%%%%%%%%%%%%%%%%%%%%%%%%%%%%%%%%%%%%%%%%%%%%%%%%%%%%%%%%%
%%%%%%%%%%%%%%%%%%%%%%%%%%%%%%%%%%%%%%%%%%%%%%%%%%%%%%%%%%%%%%%%%%%%%%%%%%%%%%%
\newpage
\appendix
\onecolumn
\input{appendix}
%%%%%%%%%%%%%%%%%%%%%%%%%%%%%%%%%%%%%%%%%%%%%%%%%%%%%%%%%%%%%%%%%%%%%%%%%%%%%%%
%%%%%%%%%%%%%%%%%%%%%%%%%%%%%%%%%%%%%%%%%%%%%%%%%%%%%%%%%%%%%%%%%%%%%%%%%%%%%%%

\end{document}

%% file: abstract.tex
We present a geometric framework for understanding Transformer-based language models, drawing an explicit analogy to General Relativity. Queries and keys induce an effective metric on representation space, and attention acts as a discrete connection that implements parallel transport of value vectors across tokens. Stacked layers provide discrete time-slices through which token representations evolve on this curved manifold, while backpropagation plays the role of a least-action principle that shapes loss-minimizing trajectories in parameter space.
If this analogy is correct, token embeddings should not traverse straight paths in feature space; instead, their layer-wise steps should bend and reorient as interactions mediated by embedding space curvature. To test this prediction, we design experiments that expose both the presence and the consequences of curvature: (i) we visualize a curvature landscape for a full paragraph, revealing how local turning angles vary across tokens and layers; (ii) we show through simulations that excess counts of sharp/flat angles and longer length-to-chord ratios are not explainable by dimensionality or chance; and (iii) inspired by Einstein’s eclipse experiment, we probe deflection under controlled context edits, demonstrating measurable, meaning-consistent bends in embedding trajectories that confirm attention-induced curvature.

%% file: analogy_table.tex
\begin{table*}[h]
\centering
\small
\renewcommand{\arraystretch}{1.3}
\begin{tabular}{|p{3.5cm}|p{5.2cm}|p{5.2cm}|}
\hline
 & \textbf{General Relativity} & \textbf{LLMs} \\
\hline
\textbf{Core Equations} & 
Einstein's Field Equations & 
Stack of Transformer layers; attention mechanism and loss minimization \\
\hline
\textbf{Curvature} & 
Geometry of spacetime & 
Geometry of embedding space \\
\hline
\textbf{Source} & 
Mass / energy distribution & 
Gradient of the loss function updates model weights \\
\hline
\textbf{Time-evolution} & 
Foliation into spacelike hypersurfaces governed by Einstein's equations & 
Sequence of Transformer blocks that progressively refine internal representations \\
\hline
\end{tabular}
\caption{Structural analogy between General Relativity and Transformer-based Language Models.}
\label{tab:gr-llm-analogy}
\end{table*}

%% file: acknowledgments.tex
\section*{Acknowledgments}
This work is dedicated to the memory of the late Prof. Silvio Bergia, who taught us Einstein's theory of General Relativity and, more importantly, the language of gravity.
%We are deeply grateful to Luis Serrano for encouraging this exploration at a crucial stage and for insightful early feedback.
A special note of appreciation to Yoshua Bengio, not for offering us a place in his lab, but for the twists and turns that lead us here. 
% And to Béatrice en baskets, l’étincelle qui a tout mis en mouvement. Je ne t’oublierai jamais.

%% file: appendix.tex
\appendix

\section*{Appendix}

\section{Introduction to the Geometry of Curved Manifolds}
\label{app:intro_curved}

In Riemannian geometry, the {\sl metric tensor} $g_{ij}$ defines local distances and angles:
\begin{equation}
    ds^2 = g_{ij}\,dx^i dx^j.
\end{equation}
If an analytical form is unavailable, the metric can often be modeled implicitly via a kernel inner product:
\begin{equation}
    ds^2 = \langle Kx, Ky \rangle.
\end{equation}
Throughout, we follow Einstein's summation convention: repeated indices imply a sum, e.g.
\begin{equation}
    A^i B_i \equiv \sum_i A^i B_i.
\end{equation}

\paragraph{Tensors and Transformations.}
A {\sl tensor} is a multilinear object whose components transform as
\begin{equation}
    T'_{ij} =
    \frac{\partial x^m}{\partial x^{\prime i}}
    \frac{\partial x^n}{\partial x^{\prime j}}
    T_{mn},
\end{equation}
under a change of coordinates $\{x^i\} \mapsto \{x^{\prime i}\}$.

\paragraph{Affine Connection.}
An {\sl affine connection} encodes how basis vectors vary across the manifold. This enables derivatives in curved spaces.  
Given two vector fields \(X\) and \(Y\), the covariant derivative \(\nabla_X Y\) measures the change of \(Y\) along \(X\). In coordinates:
\begin{equation}
    \nabla_i v^k = \partial_i v^k + \Gamma^k_{ij} v^j,
\end{equation}
where $\Gamma^k_{ij}$ are the {\sl Christoffel symbols}. For the Levi-Civita connection (metric-compatible and torsion-free),
\begin{equation}
    \Gamma^k_{ij} =
    \tfrac{1}{2} g^{kh}\!\left(
        \partial_i g_{jh} + \partial_j g_{ih} - \partial_h g_{ij}
    \right).
\end{equation}

\paragraph{Parallel Transport.}
A vector $v^i$ transported along a curve $\gamma(t)$ is said to be {\sl parallel transported} if
\begin{equation}
    \nabla_{\dot{\gamma}(t)} v(t) = 0,
\end{equation}
or equivalently,
\begin{equation}
    \frac{dv^i}{dt} + \Gamma^i_{jk} \frac{d\gamma^j}{dt} v^k = 0,
\end{equation}
where a dot denotes differentiation with respect to the affine parameter $t$.

\paragraph{Geodesics.}
A {\sl geodesic} is a curve that parallel-transports its tangent vector $v^i = \dot{\gamma}^i$:
\begin{equation}
    \frac{d^2 \gamma^i}{dt^2} +
    \Gamma^i_{\; jk} \frac{d\gamma^j}{dt} \frac{d\gamma^k}{dt} = 0.
\end{equation}

\paragraph{Curvature.}
The {\sl Riemann curvature tensor} captures how parallel transport around a loop depends on the path:
\begin{equation}
    R^i_{\; jkl} =
    \partial_k \Gamma^i_{jl} - \partial_l \Gamma^i_{jk} +
    \Gamma^i_{km} \Gamma^m_{jl} - \Gamma^i_{lm} \Gamma^m_{jk}.
\end{equation}
By contraction one obtains the {\sl Ricci tensor} $R_{jl} = R^i_{\; jil}$, and further contraction yields the {\sl scalar curvature} $R = g^{jl} R_{jl}$, a single number at each point measuring how volumes deviate from flat space.

%%%%%%%%%%%%%%%%%%%%%%%%
\section{From Attention Mechanism to Geometric Structures}
\label{app:christoffel}

This appendix provides the explicit mapping between Transformer operations and differential geometric objects introduced in Section~\ref{sec:math_foundations}.  

\subsection{Effective Metric}
Starting from the query and key projections,
\begin{equation}
    q_i = x_i W^Q, \quad k_j = x_j W^K,
\end{equation}
we define the effective metric as
\begin{equation}
    g_{ij} := q_i^\top k_j = x_i^\top (W^Q)^\top W^K x_j,
\end{equation}
with inverse $g^{ij}$ defined by $g^{ik} g_{kj} = \delta^i_j$.  

\subsection{Connection Coefficients}
In Riemannian geometry, the Christoffel symbols are given by
\begin{equation}
    \Gamma^i_{jk} = \tfrac{1}{2} g^{il}
    \left(
        \partial_j g_{kl} + \partial_k g_{jl} - \partial_l g_{jk}
    \right).
\end{equation}
Using the effective metric, we can compute its derivatives w.r.t.~token inputs $x$:
\begin{equation}
    \frac{\partial g_{ij}}{\partial x_k}
    = (W^Q)^\top W^K \, \delta_{ik} x_j
      + (W^K)^\top W^Q \, \delta_{jk} x_i.
\end{equation}

Substituting into the definition above yields the explicit expression for the Christoffel symbols in terms of $W^Q$ and $W^K$:  
\begin{equation}\label{eq:christoffel_qvk_app}
    \Gamma_{jk}^i =
    \frac{1}{2\sqrt{d}}
    \left [
    (Q^\top K)^{-1}_{ik}\left ( (W^Q)^\top W^K - (W^K)^\top W^Q 
    \right )x_j (\delta_{jk} + \delta_{kj})
    (Q^\top K)^{-1}_{il}(W^K)^\top W^Q x_l
    \right ] .
\end{equation}

Here the symmetric part governs context-dependent reshaping, while the antisymmetric part plays a role analogous to torsion.  

\subsection{Parallel Transport and Geodesics in Embedding Space}

Let $\gamma(t)$ denote the trajectory of a token representation across layers, with
\begin{equation}
    \gamma(t) \equiv x^{(t)} \in \mathbb{R}^d,
\end{equation}
where $t$ is a discrete layer index (approximating a continuous affine parameter).  

The tangent vector is then the change in embeddings across layers:
\begin{equation}
    \dot{\gamma}^i = \frac{d\gamma^i}{dt} \;\;\approx\;\; x^{(t+1)}_i - x^{(t)}_i.
\end{equation}

The parallel transport condition reads:
\begin{equation}
    \frac{dx^i}{dt} + \Gamma^i_{jk} \, \dot{\gamma}^j x^k = 0,
\end{equation}
where $\dot{\gamma}^j$ encodes how the embedding evolves through the Transformer stack.  

Substituting this into the geodesic equation gives:
\begin{equation}
    \frac{d^2 x^i}{dt^2} +
    \Gamma^i_{jk} \, \dot{\gamma}^j \dot{\gamma}^k = 0,
\end{equation}
which states that the layerwise evolution of embeddings follows a geodesic curve with respect to the effective metric $g_{ij} = q_i^\top k_j$.

Let $x^i_t$ be the token representation at layer $t$. Using finite differences for the layerwise “time”:
\begin{eqnarray}
    \dot{x}^i_t &\approx& x^i_{t+1} - x^i_t \\
    \ddot{x}^i_t &\approx& x^i_{t+1} - 2x^i_t + x^i_{t-1}.
\end{eqnarray}

Each layer has its own projections
\begin{eqnarray}
    q^{(t)}_i &=& x^{(t)}_i \, W^{Q}_{t}, \qquad
    k^{(t)}_j \;=\; x^{(t)}_j \, W^{K}_{t}, \\
    G^{(t)}_{ij} &=& \big(q^{(t)}_i\big)^\top k^{(t)}_j
                  \;=\; \big(x^{(t)}_i\big)^\top \!\big(W^{Q}_{t}\big)^\top W^{K}_{t}\, x^{(t)}_j, \qquad
    \big(G^{(t)}\big)^{-1} = \big(g^{ij}\big)^{(t)} .
\end{eqnarray}

Replacing derivatives and the connection built from $G^{(t)}$ yields
\begin{eqnarray}
    x^i_{t+1} - 2x^i_t + x^i_{t-1}
    &+& \frac{1}{2}\,\big(G^{(t)}\big)^{-1}_{i\ell}
    \Big[
        \big(\big(W^{Q}_{t}\big)^\top W^{K}_{t}\big)\,
        \big(x^j_{t+1}-x^j_t\big)\big(x^k_{t+1}-x^k_t\big)\,\delta_{j\ell} \nonumber\\
    &&\hspace{3.5cm}
        +\;\big(\big(W^{K}_{t}\big)^\top W^{Q}_{t}\big)\,
        \big(x^j_{t+1}-x^j_t\big)\big(x^k_{t+1}-x^k_t\big)\,\delta_{k\ell}
    \Big] \;=\; 0 .
\end{eqnarray}

This expression highlights that the ``acceleration'' of embeddings across layers (second finite difference) is balanced by curvature terms induced by the query and key projections, with the effective metric $(Q^\top K)$ determining how updates couple across tokens.

%%%%%%%%%%%%%%%%%%%%%%%%
\section{Least Action and Einstein’s Field Equations}
\label{app:least_action}
In general relativity, the Einstein Field Equations (EFE) are not postulated directly, but can be derived from a variational principle. Specifically, they follow from the principle of least action applied to the \textit{Einstein–Hilbert action}, which is the integral over spacetime of the Ricci scalar curvature \( R \), weighted by the metric determinant \( \sqrt{-g} \):

\[
S = \frac{1}{16\pi G} \int R \sqrt{-g} \, d^4x + S_{\text{matter}}.
\]

Here, \( G \) is Newton’s gravitational constant, and \( S_{\text{matter}} \) represents the action of any matter or energy present in the spacetime.

Varying this action with respect to the metric \( g_{\mu\nu} \) yields the Einstein Field Equations:

\[
R_{\mu\nu} - \frac{1}{2} R g_{\mu\nu} = 8\pi G \, T_{\mu\nu},
\]

where \( R_{\mu\nu} \) is the Ricci tensor, \( R \) the Ricci scalar, and \( T_{\mu\nu} \) the stress-energy tensor of matter.

This extremization principle is a deep geometrical statement: the actual geometry of spacetime is the one that extremizes the total curvature, balanced against the presence of matter. It echoes the structure we propose for language models: meaningful trajectories (e.g., in embedding space) arise from minimizing a contextual “action” shaped by loss and alignment.

\section{From Least Action to the Semantic Geodesic Equation}
\label{app:least_action_semantic}

We formalize the ``semantic least action'' used in the main text by analogy with
mechanics on a Riemannian manifold. Let $x(t)\in\mathbb{R}^d$ denote a token
representation evolving across model depth (semantic time $t$), endowed with a
metric $g_{ij}(x)$ (Sec.~\ref{sec:metric}). Consider the action
\begin{equation}
    S[x] \;=\; \int_{t_0}^{t_1}
    \bigg[
        \tfrac{1}{2}\, g_{ij}(x)\,\dot x^{\,i}\dot x^{\,j}
        \;+\; \lambda\,\mathcal{L}(x,t)
    \bigg]\,dt,
    \label{eq:semantic_action}
\end{equation}
where the kinetic term measures motion w.r.t.\ $g$, and $\mathcal{L}(x,t)$ is a
data/consistency potential (e.g.\ a local training loss); $\lambda>0$ balances the
two.

\paragraph{Euler--Lagrange and the geodesic-with-force form.}
The Lagrangian is $L(x,\dot x,t)=\tfrac12 g_{ij}(x)\dot x^i\dot x^j+\lambda\mathcal{L}(x,t)$.
The Euler--Lagrange equations
$\frac{d}{dt}\big(\partial L/\partial \dot x^i\big)-\partial L/\partial x^i=0$
give
\begin{equation}
    \frac{d}{dt}\big(g_{ij}\dot x^{\,j}\big)
    - \tfrac{1}{2}\,\partial_i g_{jk}\,\dot x^{\,j}\dot x^{\,k}
    \;+\; \lambda\,\partial_i \mathcal{L}(x,t)
    \;=\; 0.
\end{equation}
Writing this with the Levi--Civita connection
$\Gamma^{i}_{jk}=\tfrac12 g^{i\ell}(\partial_j g_{k\ell}+\partial_k g_{j\ell}-\partial_\ell g_{jk})$
yields the standard form
\begin{equation}
    \ddot x^{\,i} + \Gamma^{i}_{jk}(x)\,\dot x^{\,j}\dot x^{\,k}
    \;=\; -\,\lambda\, g^{ij}(x)\,\partial_j \mathcal{L}(x,t).
    \label{eq:forced_geodesic}
\end{equation}
When $\lambda=0$ this reduces to the geodesic equation. For $\lambda>0$,
the loss acts as a potential, generating a force whose effect is mediated by the metric, so that updates follow the geometry of the semantic manifold.

\paragraph{Discrete (layer-wise) version.}
Let layers index a discrete time $t\in\{0,\dots,L\}$ with step $\Delta t\!=\!1$ and
$x_t$ the representation at layer $t$. A second-order central difference for
$\ddot x$ and a mid-layer evaluation of Christoffel terms gives the discrete
counterpart of~\eqref{eq:forced_geodesic}:
\begin{equation}
    x_{t+1} - 2x_t + x_{t-1}
    \;+\;
    \Gamma^{i}_{jk}(x_t)\,(x_t^{\,j}-x_{t-1}^{\,j})\,(x_t^{\,k}-x_{t-1}^{\,k})
    \;=\;
    -\,\lambda\, g^{ij}(x_t)\,\partial_j \mathcal{L}(x_t,t).
    \label{eq:discrete_forced_geodesic}
\end{equation}
Rearranging gives a residual-style update:
\begin{equation}
    x_{t+1}
    \;=\;
    2x_t - x_{t-1}
    \;-\;
    \Gamma(x_t)\big[(x_t-x_{t-1}),(x_t-x_{t-1})\big]
    \;-\;
    \lambda\, g^{-1}(x_t)\,\nabla_x \mathcal{L}(x_t,t).
    \label{eq:residual_update}
\end{equation}
Thus, the depth-wise evolution comprises (i) a \emph{geometric drift} driven by
curvature via $\Gamma$, and (ii) a \emph{metric-preconditioned} descent term
$g^{-1}\nabla\mathcal{L}$, directly linking least action to the gradient
signals used in backpropagation.

\paragraph{Instantiating $g$ and $\Gamma$ with Q/K.}
Using the effective metric from attention (Sec.~\ref{sec:attention}),
\begin{equation}
    g_{ij}(x)\;=\; q_i(x)^{\!\top}\,k_j(x) \;=\; x_i^{\!\top}(W^Q)^{\!\top}W^K x_j,
    \qquad
    g^{ij} = (g_{ij})^{-1},
\end{equation}
the Christoffel symbols $\Gamma^{i}_{jk}(x)$ follow from $g$ via the Levi--Civita
formula, leading to the explicit Q/K expression derived in App.~\ref{app:christoffel}.
Substituting those into~\eqref{eq:residual_update} yields the layer update used in
the main text, making precise how attention-induced geometry (via $W^Q,W^K$) and
loss gradients jointly govern the token trajectory across layers.

\section{Experimental details}\label{app:null}

This appendix specifies detailed setup and detailed statistically confirmatory results for the curvature experiments conducted over $T$ embedding trajectories introduced in Section~\ref{exp:curvature}.  
In first experiment, each trajectory corresponds to the embedding evolution of a single randomly chosen noun or verb from one of $100$ sentences sampled from the High-Quality English sentences dataset \citep{agentlans_high_quality_english_sentences_2024}.  
In second experiment, each trajectory corresponds to a single word from the first paragraph of \emph{One Hundred Years of Solitude} \citep{marquez1967}.  
Each trajectory consists of token embeddings $\{x_1,\dots,x_L\}$ across Transformer layers.  
The key quantities of interest are the turning angles between consecutive layer steps and the overall length–to–chord elongation ratio.

\subsection{Curvature Statistics}
For each trajectory we calculate the $L{-}1$ turning angles
\[
\theta_i \;=\; \arccos
\!\left(
\frac{\Delta_i \cdot \Delta_{i+1}}
{\|\Delta_i\|\,\|\Delta_{i+1}\|}
\right), 
\qquad 
\Delta_i = x_i - x_{i-1}, 
\qquad i=1,\dots,L-1.
\]
The angles $\theta_i$ measure local curvature, while their frequency of deviation from the random orthogonality baseline ($\approx 90^\circ$) quantifies the amount of bending in embedding space.  
We summarize them by the combined tail count and the length–to–chord ratio:
\[
C \;=\; \sum_{i=1}^{L-1} \Bigl[\mathds{1}\{\theta_i < 80^\circ\} \;+\; \mathds{1}\{\theta_i > 100^\circ\}\Bigr],
\qquad
R \;=\; \frac{\sum_{i=1}^{L}\|\Delta_i\|}{\|x_L-x_0\|}.
\]
Large $C$ values indicate frequent flat and sharp directions expected from a curved space. While high $R$ values indicate globally curved or elongated trajectories.

\subsection{Construction of the Null}\label{app:construction}
For a given trajectory with step magnitudes $s_i=\|\Delta_i\|$, the null model constructs a random polyline of identical step lengths but random directions:
\begin{align}
u_i &\sim \mathrm{Unif}(\mathbb{S}^{d-1}) 
\;\;\text{(e.g., } u_i = z_i/\|z_i\|,\; z_i\sim\mathcal{N}(0,I_d)\text{)},\\
\widetilde{\Delta}_i &= s_i\,u_i,\qquad
\widetilde{x}_0 = 0,\quad \widetilde{x}_k = \sum_{i=1}^{k}\widetilde{\Delta}_i .
\end{align}
We generate $S=1000$ null draws, and for each $s=1,\dots,S$ compute
\[
\widetilde{\theta}^{(s)}_i \;=\; \arccos
\!\left(
\frac{\widetilde{\Delta}^{(s)}_i \cdot \widetilde{\Delta}^{(s)}_{i+1}}
{\|\widetilde{\Delta}^{(s)}_i\|\,\|\widetilde{\Delta}^{(s)}_{i+1}\|}
\right),
\]
and the corresponding null statistics
\[
\widetilde{C}^{(s)} \;=\; \sum_{i=1}^{L-1} \Bigl[\mathds{1}\{\widetilde{\theta}^{(s)}_i < 80^\circ\} \;+\; \mathds{1}\{\widetilde{\theta}^{(s)}_i > 100^\circ\}\Bigr],
\qquad
\widetilde{R}^{(s)} \;=\; \frac{\sum_{i=1}^{L}\|\widetilde{\Delta}^{(s)}_i\|}{\|\widetilde{x}^{(s)}_L-\widetilde{x}^{(s)}_0\|}.
\]
Fixing $\{s_i\}$ controls for layerwise step magnitudes, while randomizing directions removes learned orientation.  
Any increase of $C$ or $R$ beyond their null distributions indicates systematic curvature induced by the Transformer’s learned geometry.

\subsection{Pooled tests}
For trajectories $t=1,\dots,T$, define pooled observed statistics and pooled null draws by summing counts and averaging $R$:
\begin{align}
C_{\text{pool}}^{\text{obs}} &= \sum_{t=1}^{T} C^{(t)}, 
& \qquad \widetilde{C}_{\text{pool}}^{(s)} &= \sum_{t=1}^{T} \widetilde{C}^{(t,s)}, \\[2pt]
\bar{R}^{\text{obs}} &= \frac{1}{T}\sum_{t=1}^{T} R^{(t)}, 
& \qquad \widetilde{\bar{R}}^{(s)} &= \frac{1}{T}\sum_{t=1}^{T} \widetilde{R}^{(t,s)} .
\end{align}
Define pooled differences
\[
\Delta C_{\text{pool}}^{(s)} \;=\;  C_{\text{pool}}^{\text{obs}} - \widetilde{C}_{\text{pool}}^{(s)},
\qquad
\Delta \bar{R}^{(s)} \;=\; \bar{R}^{\text{obs}} - \widetilde{\bar{R}}^{(s)}.
\]
The pooled right–tailed $p$–values are
\begin{align}
p_{\text{MC}}\!\left(C_{\text{pool}}\right) \;=\; \frac{1+\#\{\Delta C_{\text{pool}}^{(s)} \le 0\}}{S+1},
\qquad
p_{\text{MC}}\!\left(\bar{R}\right) \;=\; \frac{1+\#\{\Delta \bar{R}^{(s)} \le 0\}}{S+1}.
\end{align}

We test whether the expected differences between observed and null statistics are negative, corresponding to curvature-induced increases in the raw statistics. Specifically,
for $p_{\text{MC}}\!\left(C_{\text{pool}}\right)$ we test
$H_0\!:\,\mathbb{E}[\Delta C_{\mathrm{pool}}] \le 0$
against
$H_1\!:\,\mathbb{E}[\Delta C_{\mathrm{pool}}] < 0$,
and for $p_{\text{MC}}\!\left(\bar{R}\right)$ we test
$H_0\!:\,\mathbb{E}[\Delta \bar{R}] \le 0$
against
$H_1\!:\,\mathbb{E}[\Delta \bar{R}] < 0$.
These one-sided alternatives correspond to the predicted direction of curvature:
larger tail counts $C$ and larger elongation ratios $R$ under learned, non-random
orientations relative to their isotropic nulls.

\subsection{Paired Mean Tests (Across Trajectories)}
Let the per–trajectory null means be
\[
\widetilde{\mu}_{C}^{(t)}=\frac{1}{S}\sum_{s=1}^{S}\widetilde{C}^{(t,s)},
\qquad
\widetilde{\mu}_{R}^{(t)}=\frac{1}{S}\sum_{s=1}^{S}\widetilde{R}^{(t,s)}.
\]
Define paired differences
\[
D_{C}^{(t)} = C^{(t)} - \widetilde{\mu}_{C}^{(t)},
\qquad
D_{R}^{(t)} = R^{(t)} - \widetilde{\mu}_{R}^{(t)}.
\]
With sample means $\bar{D}_C,\bar{D}_R$ and standard deviations $s_C,s_R$, the paired $t$–statistics and (right–tailed) $p$–values are
\begin{align}
t_{C} &= \frac{\bar{D}_{C}}{s_{C}/\sqrt{T}}, \quad p_{t}(C) = 1 - F_{t_{T-1}}(t_{C}),\\
t_{R} &= \frac{\bar{D}_{R}}{s_{R}/\sqrt{T}}, \quad p_{t}(R) = 1 - F_{t_{T-1}}(t_{R}),
\end{align}
where $F_{t_{T-1}}$ is Student’s $t$ CDF with $T{-}1$ degrees of freedom.

We test whether each trajectory’s observed statistic exceeds its own null mean,
indicating curvature beyond random orientation. For the paired differences
$D_C^{(t)} = C^{(t)} - \widetilde{\mu}_C^{(t)}$ and 
$D_R^{(t)} = R^{(t)} - \widetilde{\mu}_R^{(t)}$, 
the hypotheses are
$H_0\!:\,\mathbb{E}[D_C] \le 0$ vs.\ $H_1\!:\,\mathbb{E}[D_C] > 0$
for the combined tail count and 
$H_0\!:\,\mathbb{E}[D_R] \le 0$ vs.\ $H_1\!:\,\mathbb{E}[D_R] > 0$
for the elongation ratio. These one-sided alternatives correspond to
the expectation that curvature in learned trajectories increases both the
frequency of extreme turning angles ($C$) and the path-to-chord elongation ($R$)
relative to their trajectory-specific null baselines.

\subsection{Results}

Tables~\ref{app:sentences} and~\ref{app:paragraph} summarize the empirical results for both experiments. 
They report the average differences $\Delta C_{\text{pool}}^{(s)}$ and $\Delta \bar{R}^{(s)}$ and their corresponding pooled Monte–Carlo $p$–values ($p_{\text{MC}}$) for the extreme–angle count and length–to–chord ratio across multiple embedding models.  
For completeness, they also include the mean paired differences $\bar{D}_C,\bar{D}_R$ and their corresponding paired $t$–test $p$–values ($p_t$).  
In all cases, both $p_{\text{MC}}$ and $p_t$ fall below $0.005$, indicating that Transformer trajectories exhibit statistically significant deviations in both extreme–angle frequency and length–to–chord ratio relative to the random isotropic baseline.  
These results provide direct empirical evidence of curvature in the embedding space.

\section{Transformer models} \label{app:transformers}

We list below the Transformer models employed in our experiments, together with their corresponding Hugging Face identifiers.

\begin{itemize}
  \item \textbf{Small models:}
  \begin{itemize}
    \item DistilBERT (\texttt{distilbert-base-uncased}; \citep{sanh2019distilbert}) 
    \item DistilRoBERTa (\texttt{distilroberta-base}; \citep{sanh2019distilbert, liu2019roberta})
    \item MiniLM (\texttt{sentence-transformers/all-MiniLM-L6-v2}; \citep{reimers-2019-sentence-bert})
  \end{itemize}

  \item \textbf{Base models:}
  \begin{itemize}
    \item BERT (\texttt{bert-base-uncased}; \citep{devlin2019bert})
    \item RoBERTa (\texttt{roberta-base}; \citep{liu2019roberta})
    \item MiniLM (\texttt{sentence-transformers/all-MiniLM-L12-v2}; \citep{reimers-2019-sentence-bert})
    \item DeBERTa (\texttt{microsoft/deberta-v3-base}; \citep{he2021debertav3, he2021deberta})
    \item ALBERT (\texttt{albert-base-v2}; \citep{lan2020albert})
  \end{itemize}

  \item \textbf{Large models:}
  \begin{itemize}
    \item BERT (\texttt{bert-large-uncased}; \citep{devlin2019bert})
    \item RoBERTa (\texttt{roberta-large}; \citep{liu2019roberta})
    \item DeBERTa (\texttt{microsoft/deberta-large}; \citep{he2021deberta})
    \item ALBERT (\texttt{albert-large-v2}; \citep{lan2020albert})
  \end{itemize}
\end{itemize}

\begin{table*}[h]
\centering
\caption{
Comparison of pooled and paired statistical tests across Transformer models and sizes for first experiment on Section~\ref{sec:curvature_landscapes}.
Columns under \text{Pooled tests} report the aggregate differences in combined counts of sharp and flat angles ($\Delta C_{\mathrm{pool}}$) and mean curvature ratio ($\Delta \bar{R}$) between Transformer trajectories and their null counterparts, together with their Monte Carlo $p$-values $p_{\text{MC}}(\cdot)$.
Columns under \text{Paired t-tests} show the corresponding mean within-model differences ($\bar{D}_C$, $\bar{D}_R$) and their paired $t$-test $p$-values $p_t(\cdot)$.
All reported $p$-values are smaller than 0.005 which indicate statistically significant deviations from the null, supporting that Transformer representations follow highly curved, non-random trajectories across layers.
}
\small
\renewcommand{\arraystretch}{1.3}

\begin{tabular}{llrrrrr rrrrr}

\hline
Size & Model & \multicolumn{4}{c}{Pooled tests} & & \multicolumn{4}{c}{Paired t-tests} \\
    \cline{3-6} \cline{8-11}
     & & $\Delta C_{\mathrm{pool}}$ & $p_{\text{MC}}(C_{\mathrm{pool}})$ & $\Delta \bar{R}$ & $p_{\text{MC}}(\Delta \bar{R})$ & & $\bar{D}_C$ & $p_t(\bar{D}_C)$ & $\bar{D}_R$ & $p_t(\bar{D}_R)$ \\
     \hline

\hline 
 small & DistilBERT & 159.00 & 0.00 & 0.90 & 0.00 &  & 1.59 & 0.00 & 0.90 & 0.00 \\
 & DistilRoBERTa & 193.00 & 0.00 & 1.45 & 0.00 &  & 1.93 & 0.00 & 1.45 & 0.00 \\
 & MiniLM & 81.69 & 0.00 & 1.07 & 0.00 &  & 0.82 & 0.00 & 1.07 & 0.00 \\
 & DeBERTaV3 & 106.00 & 0.00 & -0.10 & 1.00 &  & 1.06 & 0.00 & -0.10 & 1.00 \\
\hline 
 base & BERT & 45.00 & 0.00 & 1.87 & 0.00 &  & 0.45 & 0.00 & 1.87 & 0.00 \\
 & RoBERTa & 176.00 & 0.00 & 2.79 & 0.00 &  & 1.76 & 0.00 & 2.79 & 0.00 \\
 & MiniLM & 127.31 & 0.00 & 2.30 & 0.00 &  & 1.27 & 0.00 & 2.30 & 0.00 \\
 & DeBERTaV3 & 257.00 & 0.00 & 0.08 & 0.00 &  & 2.57 & 0.00 & 0.08 & 0.00 \\
 & ALBERT & 97.00 & 0.00 & 1.07 & 0.00 &  & 0.97 & 0.00 & 1.06 & 0.00 \\
\hline 
 large & BERT & 633.00 & 0.00 & 5.35 & 0.00 &  & 6.33 & 0.00 & 5.35 & 0.00 \\
 & RoBERTa & 586.00 & 0.00 & 0.42 & 0.00 &  & 5.86 & 0.00 & 0.42 & 0.00 \\
 & DeBERTa & 143.00 & 0.00 & 1.65 & 0.00 &  & 1.43 & 0.00 & 1.65 & 0.00 \\
 & ALBERT & 655.00 & 0.00 & 1.71 & 0.00 &  & 6.55 & 0.00 & 1.71 & 0.00 \\

\hline 
\end{tabular}
\label{app:sentences}
\end{table*}

\begin{table*}[h]
\centering
\caption{
Comparison of pooled and paired statistical tests across Transformer models and sizes for second experiment on Section~\ref{sec:curvature_landscapes}.
Columns under \text{Pooled tests} report the aggregate differences in combined counts of sharp and flat angles ($\Delta C_{\mathrm{pool}}$) and mean curvature ratio ($\Delta \bar{R}$) between Transformer trajectories and their null counterparts, together with their Monte Carlo $p$-values $p_{\text{MC}}(\cdot)$.
Columns under \text{Paired t-tests} show the corresponding mean within-model differences ($\bar{D}_C$, $\bar{D}_R$) and their paired $t$-test $p$-values $p_t(\cdot)$.
All reported $p$-values are smaller than 0.005 which indicate statistically significant deviations from the null, supporting that Transformer representations follow highly curved, non-random trajectories across layers.
}
\small
\renewcommand{\arraystretch}{1.3}

\begin{tabular}{llrrrrr rrrrr}

\hline Size & Model & \multicolumn{4}{c}{Pooled tests} & & \multicolumn{4}{c}{Paired t-tests} \\
    \cline{3-6} \cline{8-11}
     & & $\Delta C_{\mathrm{pool}}$ & $p_{\text{MC}}(C_{\mathrm{pool}})$ & $\Delta \bar{R}$ & $p_{\text{MC}}(\Delta \bar{R})$ & & $\bar{D}_C$ & $p_t(\bar{D}_C)$ & $\bar{D}_R$ & $p_t(\bar{D}_R)$ \\
     \hline

\hline 
 small & DistilBERT & 196.00 & 0.00 & 1.28 & 0.00 &  & 1.59 & 0.00 & 1.28 & 0.00 \\
 & DistilRoBERTa & 205.00 & 0.00 & 1.73 & 0.00 &  & 1.67 & 0.00 & 1.73 & 0.00 \\
 & MiniLM & 128.63 & 0.00 & 2.29 & 0.00 &  & 1.05 & 0.00 & 2.29 & 0.00 \\
 & DeBERTaV3 & 123.00 & 0.00 & 0.10 & 0.00 &  & 1.00 & 0.00 & 0.10 & 0.00 \\
\hline 
 base & BERT & 161.00 & 0.00 & 2.21 & 0.00 &  & 1.31 & 0.00 & 2.21 & 0.00 \\
 & RoBERTa & 250.00 & 0.00 & 3.49 & 0.00 &  & 2.03 & 0.00 & 3.49 & 0.00 \\
 & MiniLM & 258.17 & 0.00 & 3.95 & 0.00 &  & 2.10 & 0.00 & 3.95 & 0.00 \\
 & DeBERTaV3 & 358.00 & 0.00 & 0.24 & 0.00 &  & 2.91 & 0.00 & 0.24 & 0.00 \\
 & ALBERT & 208.00 & 0.00 & 1.25 & 0.00 &  & 1.69 & 0.00 & 1.25 & 0.00 \\
\hline 
 large & BERT & 959.00 & 0.00 & 5.65 & 0.00 &  & 7.80 & 0.00 & 5.65 & 0.00 \\
 & RoBERTa & 649.00 & 0.00 & 0.57 & 0.00 &  & 5.28 & 0.00 & 0.57 & 0.00 \\
 & DeBERTa & 402.00 & 0.00 & 2.03 & 0.00 &  & 3.27 & 0.00 & 2.03 & 0.00 \\
 & ALBERT & 1419.00 & 0.00 & 3.76 & 0.00 &  & 11.54 & 0.00 & 3.76 & 0.00 \\

\hline 
\end{tabular}
\label{app:paragraph}
\end{table*}

\begin{table*}[h]
\caption{Curvature metrics of embedding trajectories from all words in the first paragraph of \emph{One hundred years of solitude} across Transformer layers.
For each model, we report counts of flat ($\theta<80^\circ$) and sharp ($\theta>100^\circ$) angles, and the average length-to-chord ratio $R$, compared to a randomized null model sampled 1000 times.}
\centering
\small
\renewcommand{\arraystretch}{1.3}

\begin{tabular}{lllrrrcrrr}
\hline
 &  &  & \multicolumn{3}{c}{\textbf{Transformers}} &  & \multicolumn{3}{c}{\textbf{Null model}} \\
\cline{4-6} \cline{8-10}
\textbf{Size} & \textbf{Model} & \textbf{Angles} & \textbf{flat} & \textbf{sharp} & \textbf{Average $R$} &  &\textbf{flat} & \textbf{sharp} & \textbf{Average $R$} \\
\hline
small & DistilBERT & 615 & 80 & 116 & 3.68 &  & 0.00 & 0.00 & 2.40 \\
 & DistilRoBERTa & 615 & 195 & 10 & 4.05 &  & 0.00 & 0.00 & 2.32 \\
 & MiniLM & 615 & 13 & 116 & 4.68 &  & 0.19 & 0.18 & 2.39 \\
 \hline
base & BERT & 1353 & 82 & 79 & 5.63 &  & 0.00 & 0.00 & 3.42 \\
 & RoBERTa & 1353 & 155 & 95 & 6.78 &  & 0.00 & 0.00 & 3.29 \\
 & MiniLM & 1353 & 144 & 115 & 7.37 &  & 0.42 & 0.40 & 3.42 \\
 & DeBERTa & 1353 & 22 & 336 & 3.02 &  & 0.00 & 0.00 & 2.79 \\
 & ALBERT & 1353 & 195 & 13 & 4.53 &  & 0.00 & 0.00 & 3.28 \\
 \hline
large & BERT & 2829 & 821 & 138 & 10.33 &  & 0.00 & 0.00 & 4.68 \\
 & RoBERTa & 2829 & 598 & 51 & 5.19 &  & 0.00 & 0.00 & 4.63 \\
 & DeBERTa & 2829 & 394 & 8 & 6.79 &  & 0.00 & 0.00 & 4.76 \\
 & ALBERT & 2829 & 258 & 1161 & 8.56 &  & 0.00 & 0.00 & 4.80 \\
\hline
\end{tabular}
\label{tab:paragraph}
\end{table*}

\begin{table*}[h]
\caption{Average trajectory divergence metrics across 50 sentence triples for experiment in Section~\ref{sec:gravitational_lensing}.
For each model, we compare the target word’s layerwise path pairwise between the \texttt{with},  \texttt{without} and \texttt{base} variants.
Reported metrics are 
$ d_{\mathrm{final}} $ (final-layer distance),
$ d_{\mathrm{layer}} $ (mean layerwise distance),
$ \Delta_{\mathrm{curv}} $ (curvature divergence), 
and $ \Delta_{\theta} $ (turning-angle gap). 
Higher values indicate stronger contextual bending. Standard deviations reported in parenthesis below each average.}
\centering
\tiny
\renewcommand{\arraystretch}{1.3}

\begin{tabular}{llrrrr r rrrr r rrrr}
\hline
\multicolumn{2}{c}{} & \multicolumn{4}{c}{\texttt{with} vs. \texttt{without}} && \multicolumn{4}{c}{\texttt{without} vs. \texttt{base}} && \multicolumn{4}{c}{\texttt{with} vs. \texttt{base}} \\
\cline{3-6}\cline{8-11}\cline{13-16}
size & model & $d_{\mathrm{final}}$ & $d_{\mathrm{layer}}$ & $\Delta_{\mathrm{curv}}$ & $\Delta_{\theta}$ & & $d_{\mathrm{final}}$ & $d_{\mathrm{layer}}$ & $\Delta_{\mathrm{curv}}$ & $\Delta_{\theta}$ & & $d_{\mathrm{final}}$ & $d_{\mathrm{layer}}$ & $\Delta_{\mathrm{curv}}$ & $\Delta_{\theta}$ \\
\hline 
 base & ALBERT & 0.15 & 10.03 & 0.16 & 0.05 &  & 0.18 & 11.19 & 0.21 & 0.05 &  & 0.05 & 4.89 & 0.07 & 0.03 \\
 &  & (0.07) & (2.58) & (0.05) & (0.01) &  & (0.09) & (3.25) & (0.09) & (0.03) &  & (0.08) & (4.04) & (0.09) & (0.04) \\
 & BERT & 0.20 & 10.22 & 0.32 & 0.05 &  & 0.26 & 11.31 & 0.39 & 0.05 &  & 0.07 & 4.47 & 0.10 & 0.02 \\
 &  & (0.10) & (2.39) & (0.12) & (0.01) &  & (0.13) & (2.32) & (0.12) & (0.01) &  & (0.12) & (2.67) & (0.11) & (0.01) \\
 & RoBERTa & 0.05 & 7.53 & 0.39 & 0.06 &  & 0.05 & 7.87 & 0.42 & 0.06 &  & 0.01 & 2.50 & 0.07 & 0.02 \\
 &  & (0.02) & (1.43) & (0.10) & (0.02) &  & (0.02) & (1.29) & (0.09) & (0.02) &  & (0.00) & (1.03) & (0.05) & (0.01) \\
 & MiniLM & 0.11 & 4.78 & 0.24 & 0.04 &  & 0.13 & 5.00 & 0.27 & 0.05 &  & 0.02 & 1.43 & 0.03 & 0.02 \\
 &  & (0.06) & (1.27) & (0.10) & (0.01) &  & (0.07) & (1.24) & (0.10) & (0.01) &  & (0.02) & (0.54) & (0.02) & (0.01) \\
\hline 
 large & ALBERT & 0.09 & 9.16 & 0.27 & 0.09 &  & 0.13 & 10.34 & 0.33 & 0.10 &  & 0.06 & 5.43 & 0.15 & 0.06 \\
 &  & (0.04) & (2.33) & (0.07) & (0.04) &  & (0.08) & (2.95) & (0.10) & (0.04) &  & (0.10) & (4.06) & (0.13) & (0.03) \\
 & BERT & 0.21 & 12.52 & 0.34 & 0.04 &  & 0.26 & 13.34 & 0.39 & 0.05 &  & 0.09 & 5.12 & 0.09 & 0.02 \\
 &  & (0.10) & (2.79) & (0.10) & (0.01) &  & (0.15) & (2.70) & (0.10) & (0.02) &  & (0.15) & (2.22) & (0.07) & (0.02) \\
 & DeBERTa & 0.28 & 23.83 & 0.52 & 0.05 &  & 0.30 & 24.80 & 0.56 & 0.05 &  & 0.06 & 8.40 & 0.12 & 0.02 \\
 &  & (0.07) & (4.36) & (0.12) & (0.01) &  & (0.07) & (3.96) & (0.10) & (0.01) &  & (0.04) & (3.43) & (0.07) & (0.01) \\
 & RoBERTa & 0.02 & 12.31 & 0.44 & 0.06 &  & 0.02 & 13.07 & 0.48 & 0.07 &  & 0.00 & 4.52 & 0.09 & 0.03 \\
 &  & (0.01) & (2.32) & (0.10) & (0.01) &  & (0.01) & (1.95) & (0.08) & (0.01) &  & (0.00) & (1.82) & (0.07) & (0.01) \\
\hline 
 small & DistilBERT & 0.12 & 6.84 & 0.24 & 0.06 &  & 0.16 & 7.52 & 0.30 & 0.06 &  & 0.03 & 2.91 & 0.06 & 0.02 \\
 &  & (0.05) & (1.45) & (0.08) & (0.03) &  & (0.06) & (1.45) & (0.08) & (0.02) &  & (0.04) & (1.44) & (0.06) & (0.02) \\
 & DistilRoBERTa & 0.04 & 6.05 & 0.39 & 0.09 &  & 0.04 & 6.34 & 0.42 & 0.09 &  & 0.00 & 1.70 & 0.05 & 0.02 \\
 &  & (0.02) & (1.26) & (0.11) & (0.03) &  & (0.02) & (1.05) & (0.09) & (0.04) &  & (0.00) & (0.81) & (0.04) & (0.01) \\
 & DeBERTaV3 & 0.18 & 12.20 & 0.44 & 0.05 &  & 0.17 & 12.71 & 0.48 & 0.05 &  & 0.03 & 3.73 & 0.07 & 0.02 \\
 &  & (0.09) & (2.38) & (0.11) & (0.02) &  & (0.07) & (1.97) & (0.09) & (0.02) &  & (0.02) & (2.09) & (0.07) & (0.01) \\
 & MiniLM & 0.11 & 4.44 & 0.22 & 0.04 &  & 0.13 & 4.61 & 0.25 & 0.04 &  & 0.02 & 1.35 & 0.03 & 0.01 \\
 &  & (0.06) & (1.02) & (0.09) & (0.02) &  & (0.07) & (1.00) & (0.09) & (0.02) &  & (0.02) & (0.56) & (0.02) & (0.01) \\
\hline
\end{tabular}

\label{tab:sun}
\end{table*}